\documentclass[lettersize,journal]{IEEEtran}
\usepackage{amsmath,amsfonts}
\usepackage{algorithmicx}
\usepackage{algorithm}
\usepackage{array}
\usepackage[caption=false,font=normalsize,labelfont=sf,textfont=sf]{subfig}
\usepackage{textcomp}
\usepackage{stfloats}
\usepackage{amssymb}
\usepackage{algpseudocode}
\usepackage{enumitem}
\usepackage{url}
\usepackage{verbatim}
\usepackage{graphicx}
\usepackage{booktabs}
\usepackage{subfig}   % 推荐使用 subfig
\usepackage{cite}
\usepackage{threeparttable}
\usepackage{float}           % 提供 [H] 选项
\usepackage{booktabs}        % 提供 \toprule, \midrule, \bottomrule, \addlinespace
\usepackage{tabularx}        % 提供 tabularx 环境
\usepackage{bm}
\begin{document}

\title{Graph Embedding in the Graph Fractional \\Fourier Transform Domain}

\author{Changjie~Sheng, Zhichao~Zhang,~\IEEEmembership{Member,~IEEE}, and Yangfan~He
	\thanks{This work was supported in part by the Open Foundation of Hubei Key Laboratory of Applied Mathematics (Hubei University) under Grant HBAM202404; in part by the Foundation of Key Laboratory of System Control and Information Processing, Ministry of Education under Grant Scip20240121; and in part by the Startup Foundation for Introducing Talent of Nanjing Institute of Technology under Grant YKJ202214. \emph{(Corresponding author: Zhichao~Zhang.)}}
	\thanks{Changjie~Sheng is with the School of Mathematics and Statistics, Nanjing University of Information Science and Technology, Nanjing 210044, China (e-mail: scj135799@163.com).}
	\thanks{Zhichao~Zhang is with the School of Mathematics and Statistics, Nanjing University of Information Science and Technology, Nanjing 210044, China, with the Hubei Key Laboratory of Applied Mathematics, Hubei University, Wuhan 430062, China, and also with the Key Laboratory of System Control and Information Processing, Ministry of Education, Shanghai Jiao Tong University, Shanghai 200240, China (e-mail: zzc910731@163.com).}
	\thanks{Yangfan~He is with the School of Communication and Artificial Intelligence, School of Integrated Circuits, Nanjing Institute of Technology, Nanjing 211167, China, and also with the Jiangsu Province Engineering Research Center of IntelliSense Technology and System, Nanjing 211167, China (e-mail: Yangfan.He@njit.edu.cn).}}



\maketitle

\begin{abstract}
	Spectral graph embedding plays a critical role in graph representation learning by generating low-dimensional vector representations from graph spectral information. However, the embedding space of traditional spectral embedding methods often exhibit limited expressiveness, failing to exhaustively capture latent structural features across alternative transform domains. To address this issue, we use the graph fractional Fourier transform to extend the existing state-of-the-art generalized frequency filtering embedding (GEFFE) into fractional domains, giving birth to the generalized fractional filtering embedding (GEFRFE), which enhances embedding informativeness via the graph fractional domain.The GEFRFE leverages graph fractional domain filtering and a nonlinear composition of eigenvector components derived from a fractionalized graph Laplacian. To dynamically determine the fractional order, two parallel strategies are introduced: search-based optimization and a ResNet18-based adaptive learning. Extensive experiments on five benchmark datasets demonstrate that the GEFRFE captures richer structural features and significantly enhance classification performance. The GEFRFE provides a new paradigm for the development of graph embedding from the "fixed domain" to the "generalized domain". The results indicate that introducing the GFRFT into the graph embedding domain is a correct and effective research path. Notably, the proposed method retains computational complexity comparable to GEFFE approaches.
\end{abstract}

\begin{IEEEkeywords}
	Spectral graph embedding, graph fractional Fourier transform, graph classification, graph fractional filtering.
\end{IEEEkeywords}

\section{Introduction}\label{sec:introduction}
\IEEEPARstart{G}{raph} structured data arise naturally in various domains, including social networks, biological systems, and transportation infrastructures \cite{ortega2018graph,sandryhaila2014big}. Unlike traditional data formats such as time series or images, graphs exhibit complex relational structures, posing challenges for downstream tasks such as classification, clustering, and prediction. To mitigate these challenges, graph embedding techniques are employed to convert high-dimensional, sparse graph representations into low-dimensional, dense vector spaces suitable for machine processing. Thus, the processing and modeling requirements of computers are met. Therefore, graph embedding is an indispensable technology in graph data mining.

The core requirement of graph embedding is the preservation of structural and semantic information from the original graph. This paper focuses on the research of spectral graph embedding methods. For other typologies of graph embedding methods, the literatures \cite{wang2017knowledge,cai2018comprehensive,perozzi2014deepwalk,wu2023survey,li2021learning,8937811,9741755} are quite useful. Spectral graph embedding, as one of the most fundamental and widely applied graph embedding techniques, originated from the development of spectral graph theory\cite{deng2019graphzoom,luo2003spectral}. Among the numerous embedding paradigms, spectral graph embedding—rooted in spectral graph theory—has shown substantial promise. This approach characterizes graph properties through the spectral decomposition of adjacency or Laplacian matrices  \cite{chung1997spectral,emms2005matrix}.

Early work such as \cite{xiao2009graph} introduced the heat kernel trace method, which relies on the spectral properties of the Laplacian. The heat kernel is obtained by solving the heat equation on the graph, and the moment invariants of the heat kernel trace are applied to describe the graph structure. Since the power series expansion of heat content requires the calculation of high-order terms of eigenvalues and eigenvectors, it is less efficient for larger or dense graphs. Similarly, \cite{wilson2014graph} utilized thermal diffusion to construct the heat kernel signature, focusing on node-level features, whereas \cite{xiao2009graph} emphasized graph-level embedding.

In \cite{ren2010graph}, the Ihara coefficient was proposed to describe the graph through the method of spectral analysis. This coefficient, derived from the Perron-Frobenius operator on the associated oriented line graph, captures connectivity information via the quasi-characteristic polynomial. These coefficients encode the spectral signature of a graph and implicitly capture information on graph connectivity patterns, thereby achieving the embedding and clustering analysis of the graph. Despite its theoretical strength, the method suffers from coefficient redundancy, which can degrade embedding performance.

The work in \cite{bahonar2019graph} pioneered the integration of the graph Fourier transform (GFT) into the embedding framework, leading to the development of generalized frequency filtering embedding (GEFFE). This method combines the symmetric polynomial functions proposed in \cite{wilson2005pattern}, exhaustively mines the information in the feature vector elements, and combines the filtering operation in the graph spectral domain. It can be noted that the extraction and embedding of graph structure features have been achieved. The experimental results indicate that, among the existing spectral embedding techniques \cite{aziz2013backtrackless,gartner2003graph,ren2010graph,xiao2009graph}, the GEFFE method performs excellently. The various spectral graph embedding methods essentially complete feature extraction by performing different forms of operations on the eigenvalues (frequencies) and eigenvector elements of the graph Laplacian matrix. Therefore, they can all be regarded as embeddings implemented in the graph spectral domain. Nonetheless, spectral embedding methods such as GEFFE are constrained to the spectral domain, limiting their capacity to capture information inherent in alternative transform domains.

To overcome these limitations, the graph fractional Fourier transform (GFRFT) becomes a focus of investigation. The GFRFT is an extension of the GFT. Like the fractional Fourier transform, the GFRFT enables signal analysis across a continuum of domains between the vertex and spectral domains. While earlier formulations restricted the fractional order to finite intervals  \cite{wang2017fractional}, later work generalized it to the full real line using the hyper-differential operator \cite{alikacsifouglu2024graph}.  Furthermore, theories such as sampling in the graph fraction domain and optimal filtering are also gradually being improved \cite{wang2018fractional,wei2024generalized,ozturk2021optimal,yan2021windowed,yan2020windowed,gan2025windowed,zhang2025graph,yan2023spectral,yan2022multi,wei2024hermitian}.

Therefore, the main motivation of this paper is: to solve the single problem of the embedding space and insufficient graph representational capacity by using the fractional Fourier transform of the graph, with the aim of exploring the information of the graph in other continuous transform domains, so that the feature information after embedding is as much as possible. We will support our idea from the following two points:
\begin{itemize}
	\item	Theoretically: The GFRFT matrix is essentially the fractional transform of the GFT matrix, which is equivalent to the fractionization of eigenvectors, that is, the basis fractionization. Using the fractionalized basis to represent the graph structure has a stronger and more flexible graph representation ability.
	\item	Spatially: Since the graph spectral domain space is only a special case of the graph fractional domain space. The graph fractional domain subsumes both vertex and spectral domains, offering a larger feature space that reveals deeper structural information.
\end{itemize}
The primary contributions of this work are summarized as follows:
\begin{itemize}
	\item	The graph embedding form of the graph fractional domain is defined, named the generalized fractional filtering embedding (GEFRFE), which integrates graph fractional domain filtering and power-based combinations of fractionalized eigenvector elements. This approach extends the existing GEFFE framework and is designed to overcome the limitations of a singular embedding space.
	\item	Regarding the selection problem of the fractional order of the GFRFT, this paper provides two parallel schemes: the grid search and the adaptive learning. The grid search is combined with KNN classifier to complete the classification task. The adaptive learning strategy selects ResNet-18 to achieve adaptive learning of the fractional order and accomplish the final classification task.
	\item  In the experimental section, the GEFRFE method is employed to conduct graph embedding and classification tasks on five real-world datasets. Results consistently show superior performance compared to the existing method. In addition, the GEFRFE method is consistent with the GEFFE method in terms of computational complexity.
\end{itemize}

This paper is structured through the following steps: Section \ref{Preliminaries} provides a brief introduction to the fundamental concepts of graph signals, GFRFT, and graph filtering, along with the origin of the filtering functions employed in this work. Section \ref{method} describes the proposed graph embedding method based on the graph fractional domain. Section \ref{algorithm} introduces two strategies for selecting the fractional order of the GFRFT.  Section \ref{EXPERIMENTAL} presents the experimental results and provides a computational complexity analysis comparing existing methods with the proposed approach. Finally, the paper concludes with a summary and discusses potential directions for future research.

\section{Preliminaries }\label{Preliminaries}
\subsection{Graph Signals }
Considering a graph $\mathcal{G} = \{\mathcal{V}, \mathbf{A}\}$, where $\mathcal{V} = [v[1],\dots, v[N] ]$ is the set of vertices and $\mathbf{A}$ is the adjacency matrix of the graph $\mathcal{G}$. The adjacency matrix $\mathbf{A} \in \mathbb{C}^{N \times N}$ presents the relationship between the vertices. If there is an edge from node $n$ to node $m$ then $\mathbf{A}_{n,m} \neq 0$, and $\mathbf{A}_{n,m} = 0$ otherwise, where $\mathbf{A}_{n,m}$ is the element at $n$th row and $m$th column of $\mathbf{A}$. If the graph is undirected we have $\mathbf{A}_{n,m} = \mathbf{A}_{m,n}$.
For a graph signal $\mathbf{x} = [x[1],x[2],\dots,x[N]]^{\mathsf{T}}$, $\mathbf{x} \in \mathbb{C}^{N}$, be a mapping between the set of vertices  $\mathcal{V}$ to $\mathbb{C}^{N}$ so
that each vertex is mapped to a complex number as $ \mathbf{x} : \mathcal{V} \rightarrow \mathbb{C} $ such that $v[n] \rightarrow x[n]$ \cite{6409473}. 

In this paper, we only consider the simple and undirected graph, so the adjacency matrix can be written as 
\begin{equation}\label{eq1}
	\mathbf{A}_{n, m} =
	\begin{cases}
		1 & \text{if } (n, m) \in \mathbf{E} \\
		0 & \text{otherwise}, 
	\end{cases}
\end{equation} where $\mathbf{E} \subseteq \mathcal{V} \times \mathcal{V}$ is the set of edges. The degree matrix $\mathbf{D}$ is constructed by placing the column sums of the adjacency matrix as diagonal
elements, while setting the off-diagonal elements to zero. The graph Laplacian matrix is given by $\mathbf{L} = \mathbf{D} - \mathbf{A}$.

\subsection{The Definition of GFRFT}
Let the graph Laplacian matrix $\mathbf{L}$
represent a graph shift operator matrix of an arbitrary graph $\mathcal{G}$, which can be expressed in its Jordan decomposition as $\mathbf{L} = \mathbf{V}\mathbf{J}_{\mathbf{L}}\mathbf{V}^{-1}$. In this case, the GFT matrix is defined as $\mathbf{F} = \mathbf{V}^{-1} = \mathbf{P}\mathbf{J}_{\mathbf{F}}\mathbf{P}^{-1}$, where $\mathbf{J}_{\mathbf{F}}$ is the Jordan form of $\mathbf{F}$ and $\mathbf{P}$ is the corresponding matrix whose columns contain the generalized
eigenvectors of $\mathbf{F}$. Thus, the GFRFT matrix with fractional order $ \alpha $ is given by:
\begin{equation}\label{eq2}
	\mathbf{F}^{\alpha} = \mathbf{P}\mathbf{J}^{\alpha}_{\mathbf{F}}\mathbf{P}^{-1}.
\end{equation}
The computation details of Eq. \eqref{eq2} can be found in \cite{wang2017fractional}. Then, the definition of the $\alpha$th order GFRFT is given by
\begin{equation}\label{eq3}
	\mathbf{\hat{x}}^{\alpha} = \mathbf{F}^{\alpha}\\ 
	\mathbf{x} 
\end{equation} or 
\begin{equation}\label{eq3.1}
	\mathbf{\hat{x}}^{\alpha}[\lambda_{l}] = \sum_{n=1}^{N}x[n]\Phi_l^{\alpha}[n],
\end{equation} where $\lambda_{l}$ is the $l$th eigenvalue of the Laplacian matrix $\mathbf{L}$, $\Phi_l^{\alpha}$ is the $l$th column vector of the matrix $\mathbf{F}^{\alpha}$.
The inverse GFRFT is specified by the matrix $\mathbf{F}^{-\alpha}$. Clearly,
when $\alpha = 0$ and $\alpha = 1$, the GFRFT reduces to $\mathbf{I}$ (the identity
matrix) and GFT, respectively. Further details of the GFRFT for properties can be found in \cite{wang2017fractional} and \cite{wang2018fractional}.

Since the Laplacian matrix $\mathbf{L}$ is real and symmetric, it is always diagonalizable. Therefore, all matrix decompositions used in this work are based on eigendecomposition, without the need for the more general Jordan decomposition.

Therefore, we introduce the fractional Fourier transform for graph signals, which will serve as the foundation for the following sections.

\subsection{Graph Filters and Filter Function}
Frequency filtering is an operator in the  frequency domain which amplifies or attenuates some frequencies. Similarly, the GFRFT domain filtering is defined as:
\begin{equation}\label{eq4}
	\mathbf{\hat{x}}_{out}^{\alpha} = \mathbf{H\hat{x}}^{\alpha}
\end{equation} 
or
\begin{equation}\label{eq4.4}
	\mathbf{\hat{x}}_{out}^{\alpha}[\lambda_{l}] = H[\lambda_{l}]\mathbf{\hat{x}}^{\alpha}[\lambda_{l}],
\end{equation}
where $\mathbf{H} $ is the diagonal matrix of filter function whose inputs are eigenvalues, $H$ is the filter function. 

In \cite{shuman2013emerging}, there is a relation between heat kernel and GFT. Assume $X$ as the initial heat on the graph nodes. The heat amount on each
node $x[n]$ at time $t$, $H_{t}X[n]$ can be found by
\begin{equation}\label{eq5}
	H_tX[n] = \sum_{m=1}^{N} K_t[n,m]x[m],
\end{equation} where 
\begin{equation}\label{eq6}
	K_t[n,m] = \sum_{l=1}^{N} e^{-\lambda_l t} \phi_l[n]\phi_l[m]
\end{equation} is the transmitted heat from node $n$ to node $m$ at time $t$ \cite{bai2007heat}, $\phi_l$ is the $l$th eigenvector of the Laplacian matrix $\mathbf{L}$. By substituting Eq. \eqref{eq3} into Eq. \eqref{eq5} yields 
\begin{align}\label{eq8}
	H_tX[n] &= \sum_{m=1}^{N}\sum_{l=1}^{N} e^{-\lambda_l t} \phi_l[n]\phi_l[m] x[m]\\
	&= \sum_{l=1}^{N} e^{-\lambda_l t}\hat{\mathbf{x}}[\lambda_{l}] \phi_l[n] = \sum_{l=1}^{N} \widehat{H_tX[\lambda_{l}]}\phi_l[n],\nonumber
\end{align} 
where $\hat{\mathbf{x}}$ denotes the GFT of $ \mathbf{x}$.  Comparing Eq. \eqref{eq4.4} and Eq. \eqref{eq8}, $H_tX$ is the result of frequency of filtering on $\mathbf{x}$ by filter function $\hat{H}_{t}[\lambda_{l}] = e^{-\lambda_{l}t}$. In the GFRFT domain, we can still employ this filter function.

\section{Graph Embedding in GFRFT Domain}\label{method}
In this section, we will introduce the graph embedding using filtering in the GFRFT domain.

For a constant graph signal, i.e., $x[n] =1, n \in \{1,2\dots,N\}$, the GFRFT is given by $\mathbf{\hat{x}}^{\alpha}[\lambda_{l}] = \sum_{n=1}^{N}\Phi_l^{\alpha}[n]$, which can be regards as  the summation of each element of the eigenvector. Moreover, in order to reduce the influence of cancellation when summing each eigenvector element, the $\hat{x}$ can be rewritten as $\mathbf{\hat{x}}^{\alpha}_{\omega}[\lambda_{l}] = \sum_{n=1}^{N}\left[\Phi^{\alpha}_l[n]\right]^{\omega}$. Therefore, we have the following definition.

\textbf{Definition 1}. Generalized fractional filtering embedding (GEFRFE): Let $\mathcal{H} = \{H_1, H_2, \dots, H_r\}$ be the filter bank, where $H_k$, $ \mathbb{R} \rightarrow \mathbb{R}$, $k \in \{1,2,\dots, r\} $ is a filter function, $\mathcal{W} = \{\omega_{1},\omega_{2},\dots,\omega_{\Omega}\}$. The GEFRFE $\mathcal{E}_{\mathcal{H}}$ is defined as:
\begin{align}\nonumber\label{eq13}
	\mathcal{E}_{\mathcal{H}}^{\alpha}: \mathcal{G} \rightarrow &\mathbb{R}^{N \times r} \\\nonumber
	\mathcal{E}_{\mathcal{H}}^{\alpha}[\mathbf{x}] =
	\Bigl[&\mathbf{f}^{\alpha}_{1,\omega_{1}}[\mathbf{x}]^{\mathsf{T}}, \mathbf{f}^{\alpha}_{1,\omega_{2}}[\mathbf{x}]^{\mathsf{T}}, \dots, \mathbf{f}^{\alpha}_{1,\omega_{\Omega}}[\mathbf{x}]^{\mathsf{T}},\\
	&\mathbf{f}^{\alpha}_{2,\omega_{1}}[\mathbf{x}]^{\mathsf{T}}, \mathbf{f}^{\alpha}_{2,\omega_{2}}[\mathbf{x}]^{\mathsf{T}}, \dots, \mathbf{f}^{\alpha}_{2,\omega_{\Omega}}[\mathbf{x}]^{\mathsf{T}},\\\nonumber
	&\vdots\\
	&\mathbf{f}^{\alpha}_{r,\omega_{1}}[\mathbf{x}]^{\mathsf{T}}, \mathbf{f}^{\alpha}_{r,\omega_{2}}[\mathbf{x}]^{\mathsf{T}}, \dots, \mathbf{f}^{\alpha}_{r,\omega_{\Omega}}[\mathbf{x}]^{\mathsf{T}}\Bigr],\nonumber
\end{align} where $\mathbf{f}^{\alpha}_{k,\omega_j}[\mathbf{x}] = H_{k}[\lambda_{l}]\hat{\mathbf{X}}_{\omega_{j}}^{\alpha}[\lambda_{l}]$, $l \in \{1,2,\dots,N\}$ is the response of graph signal $\mathbf{x}$ in  power order $\omega$ to the filter $H_{k}$, and where $\hat{\mathbf{X}}^{\alpha}_{\omega}$ is defined as:
\begin{equation}\label{eq14}
	\hat{\mathbf{X}}^{\alpha}_{\omega} = \left\{\sum_{u=1}^{N}\left[\Phi_{1}^{\alpha}[u]\right]^{\omega},\sum_{u=1}^{N}\left[\Phi_{2}^{\alpha}[u]\right]^{\omega}, \dots, \sum_{u=1}^{N}\left[\Phi_{N}^{\alpha}[u]\right]^{\omega}\right\}. 
\end{equation} \\ 
%For eq. \eqref{eq14},  when $(\alpha, \omega) = (1,1)$, $	\hat{\mathbf{X}}^{\alpha}_{\omega}$ is the GFT which conveys the notion of frequency and $\mathcal{E}_{\mathcal{H}}^{\alpha}[x]$ reduces to the frequency filtering embedding, and when $(\alpha,\omega ) = (1,\mathbb{R})$, $\mathcal{E}_{\mathcal{H}}^{\alpha}$ reduces to the generalized frequency filtering embedding (GEFFE). 

\begin{figure*}[htb] 
	\centering
	\subfloat[] {
		\includegraphics[height=2.7cm,width=2.7cm]{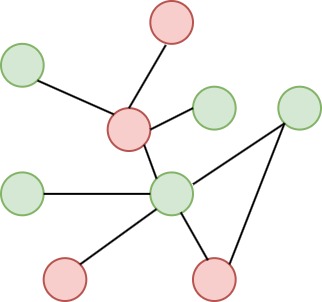}
	}
	\subfloat[] {
		\includegraphics[height=2.7cm,width=2.7cm]{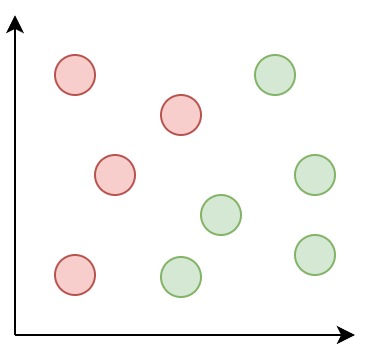}
	}
	\subfloat[] {
		\includegraphics[height=2.7cm,width=2.7cm]{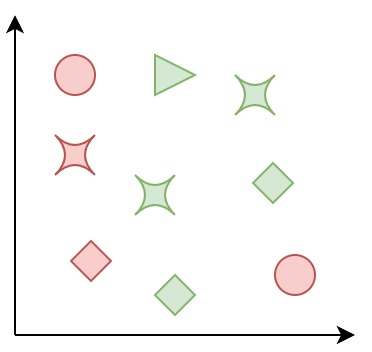}
	}
	\subfloat[] {
		\includegraphics[height=2.7cm,width=2.7cm]{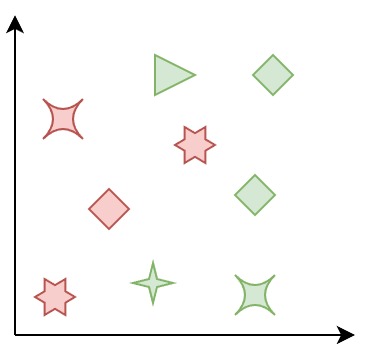}
	}
	\raisebox{1.2cm}{\makebox[1cm]{\dots}}
	\subfloat[] {
		\includegraphics[height=2.7cm,width=2.7cm]{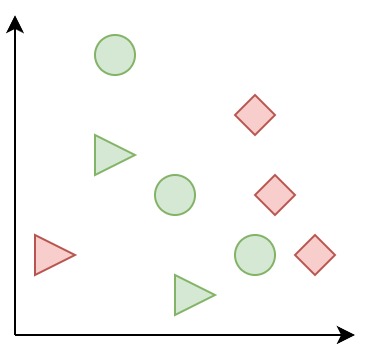}
	}
	
	\caption{The structural information of a graph in different embedding domains. (a): Vertex domain. (b): Graph spectral domain embedding, $\alpha = 1$.  (c): GFRFT domain embedding, $\alpha = \alpha_1$. (d): GFRFT domain embedding, $\alpha = \alpha_2$. (e): GFRFT domain embedding, $\alpha = \alpha_i$, $\alpha_i \in \mathbb{R} $. The transition from (b) to (e) reflects the distinct structural information encoded in the graph signal under different transform domains embedding. Circles, diamonds, and triangles are used to represent different types of structural information.}
	\label{differents}
\end{figure*}

\begin{figure*}[htb]  
	\centering
	\subfloat[]{
		\centering
		\includegraphics[height=2.3cm,width=2.3cm]{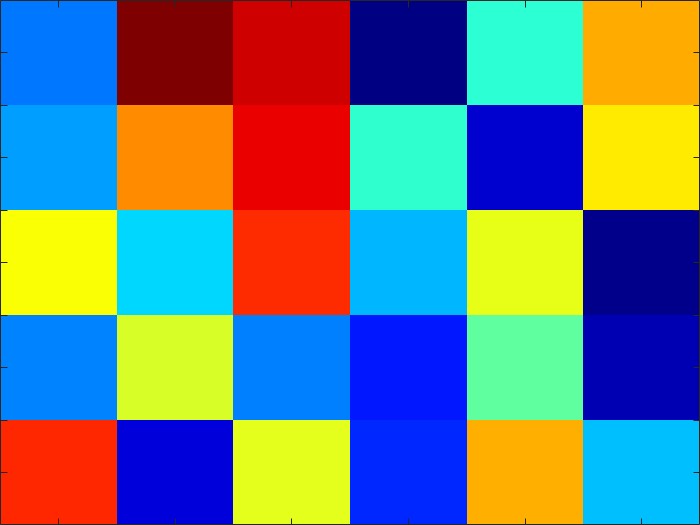}
	}
	\hspace{0.4cm}
	\subfloat[]{
		\centering
		\includegraphics[height=2.3cm,width=2.3cm]{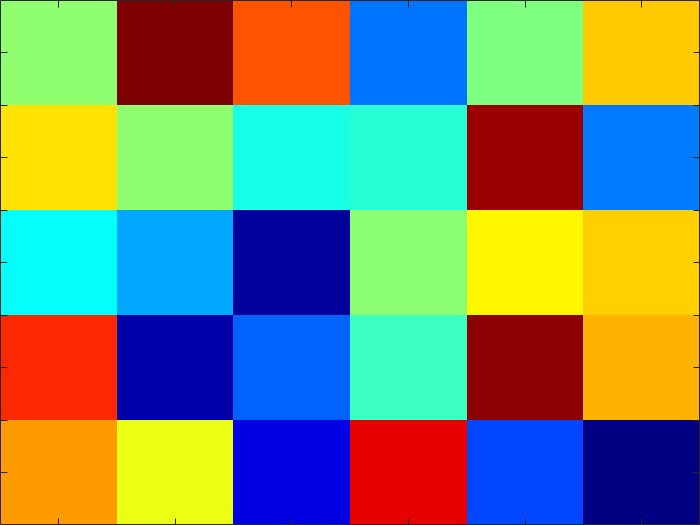}
	}
	\hspace{0.4cm}
	\subfloat[]{
		\centering
		\includegraphics[height=2.3cm,width=2.3cm]{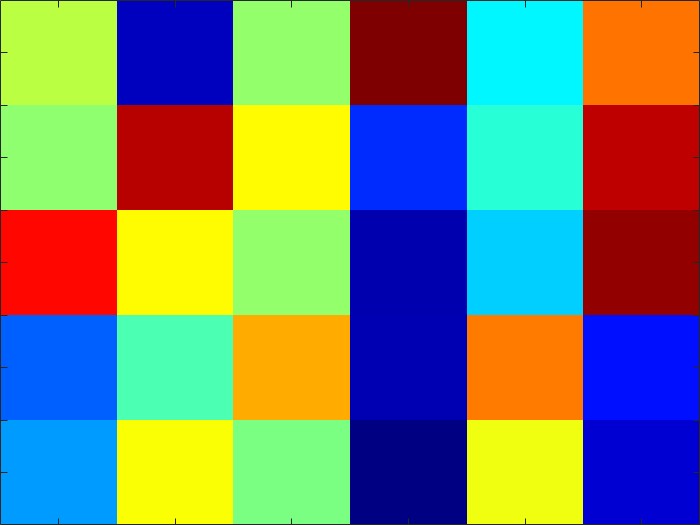}
	}
	\hspace{0.4cm}
	\raisebox{0.9cm}{\makebox[1cm]{\dots}}
	\hspace{0.4cm}
	\subfloat[]{
		\centering
		\includegraphics[height=2.3cm,width=2.3cm]{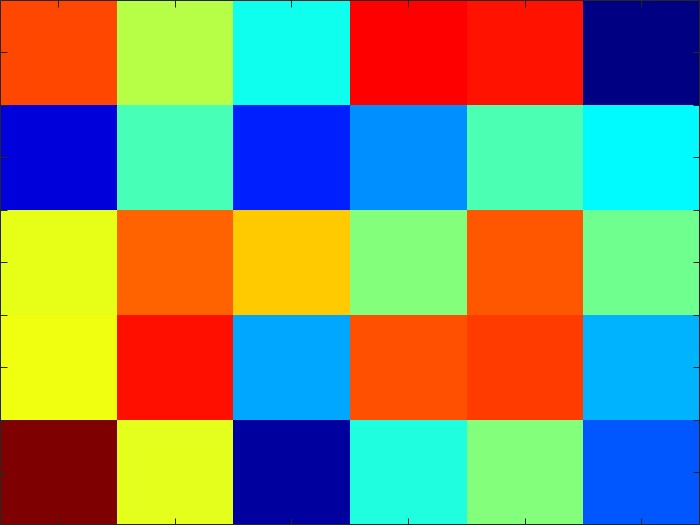}
	}
	\caption{The Visualization schematic diagram of the embedded feature matrix of multiple graph signals. 
		The rows of the matrix represent different graph signals, and every element corresponds to $\mathbf{f}^{\alpha}_{k,\omega_j}$ in Eq. \eqref{eq13}. (a): The embedding feature matrix using the method of GEFFE, $\alpha = 1 $. (b)(c)(d): The embedding feature matrix is generated using the GEFRFE method. The fractional orders corresponding to (b) - (d) are $\alpha_1$, $\alpha_2$, and $\alpha_i$, $\alpha_i \in \mathbb{R} $ in sequence. (b) to (d) indicate that the embedding feature matrix varies with the fractional order $\alpha$. }
	\label{matrix}
\end{figure*}

\textit{Comparisons with the GEFFE}: The GEFFE operates in the graph spectral domain, with its primary limitation being a single embedding space, which fails to capture the spatial structural information of graph signals in graph fractional domains. In contrast, the method GEFRFE extends the embedding space from the graph spectral domain space to the graph fractional domain space, potentially enabling deeper exploration of latent structural features in the graph. Fig. \ref{differents} plots the structural information of a graph in the  different transform domain. A schematic visualization of the embedding feature matrix for multiple graph signals is presented in Fig.~\ref{matrix}. For Eq. \eqref{eq14},  when $(\alpha, \omega) = (1,1)$, $	\hat{\mathbf{X}}^{\alpha}_{\omega}$ is the GFT which conveys the notion of frequency and $\mathcal{E}_{\mathcal{H}}^{\alpha}[x]$ reduces to the frequency filtering embedding, and when $(\alpha,\omega ) = (1,\mathbb{R})$, $\mathcal{E}_{\mathcal{H}}^{\alpha}$ reduces to the GEFFE, given by:

\begin{align}\nonumber\label{eq13.5}
	\mathcal{E}_{\mathcal{H}}: \mathcal{G} \rightarrow &\mathbb{R}^{N \times r} \\\nonumber
	\mathcal{E}_{\mathcal{H}}[\mathbf{x}] =
	\Bigl[&\mathbf{f}_{1,\omega_{1}}[\mathbf{x}]^{\mathsf{T}}, \mathbf{f}_{1,\omega_{2}}[\mathbf{x}]^{\mathsf{T}}, \dots, \mathbf{f}_{1,\omega_{\Omega}}[\mathbf{x}]^{\mathsf{T}},\\
	&\mathbf{f}_{2,\omega_{1}}[\mathbf{x}]^{\mathsf{T}}, \mathbf{f}_{2,\omega_{2}}[\mathbf{x}]^{\mathsf{T}}, \dots, \mathbf{f}_{2,\omega_{\Omega}}[\mathbf{x}]^{\mathsf{T}},\\\nonumber
	&\vdots\\
	&\mathbf{f}_{r,\omega_{1}}[\mathbf{x}]^{\mathsf{T}}, \mathbf{f}_{r,\omega_{2}}[\mathbf{x}]^{\mathsf{T}}, \dots, \mathbf{f}_{r,\omega_{\Omega}}[\mathbf{x}]^{\mathsf{T}}\Bigr],\nonumber
\end{align} where $\mathbf{f}_{k,\omega_j}[\mathbf{x}] = H_{k}[\lambda_{l}]\hat{\mathbf{X}}_{\omega_{j}}[\lambda_{l}]$, and where $\hat{\mathbf{X}}_{\omega}$ is defined as:
\begin{equation}\label{eq14.5}
	\hat{\mathbf{X}}_{\omega} = \left\{\sum_{u=1}^{N}\left[\Phi_{1}[u]\right]^{\omega},\sum_{u=1}^{N}\left[\Phi_{2}[u]\right]^{\omega}, \dots, \sum_{u=1}^{N}\left[\Phi_{N}[u]\right]^{\omega}\right\}. 
\end{equation} 

\section{Algorithms for Selecting Fractional Order $\alpha$ } \label{algorithm}
In this section, we illustrate in detail two  strategies for selecting fractional order $\alpha$, the grid search strategy and the adaptive learning strategy. These two approaches are presented in parallel, and no direct comparison is made between them. The grid search strategy will be employed in all subsequent experiments, whereas the adaptive learning strategy is applied only in the final experiment.
\subsection{The Grid Search Strategy}
We adopt the grid search to control fractional order $\alpha$ in the range $(-3, 3)$ and the search interval is $\Delta\alpha=$ 0.02. The selection of both the search range and the search interval is guided by considerations of computational time efficiency. The core idea of this strategy is to iterate over each value of the fractional order $\alpha$, compute the corresponding GEFRFE, and subsequently obtain the classification accuracy associated with each fractional order $\alpha$. The classification accuracy is assessed using a 5-nearest neighbors (5NN) classifier with 5-fold cross-validation, and the average performance across 20 runs is reported. The detailed steps by the following  \textbf{Algorithm \ref{al1}}. 
\begin{algorithm}[htb]
	\caption{Change fractional order $\alpha$ by the grid search method}
	\begin{algorithmic}[1]
		\renewcommand{\algorithmicrequire}{\textbf{Input:}}
		\renewcommand{\algorithmicensure}{\textbf{Output:}}
		\Require Adjacency matrices $\mathbf{A}$, labels, search range $(-3, 3)$, the search interval $\Delta\alpha = 0.02$, some power orders $\omega$, some filters $H_k$
		\Ensure Classification accuracy
		\State  Compute the Laplacian matrices $\mathbf{L}$: $\mathbf{L} = \mathbf{D} - \mathbf{A}$
		\For{$\alpha_i = -3$ \textbf{to} $3$}
		\State  $\mathbf{L} = \mathbf{V} \mathbf{J}_{\mathbf{L}} \mathbf{V}^{-1}$,
		$\mathbf{F}^{\alpha_i} = \mathbf{P} \mathbf{J}^{\alpha_i}_{\mathbf{F}} \mathbf{P}^{-1}$ and 
		$\hat{\mathbf{x}}^{\alpha_i} = \mathbf{F}^{\alpha_i} \mathbf{x}$
		\For{each power order $\omega_{j}$}
		\State $\mathbf{f}^{\alpha_i}_{k,\omega_j} = H_k[\lambda_l] \hat{\mathbf{X}}^{\alpha_i}_{\omega_j}[\lambda_l]$,  and $\mathcal{E}_{\mathcal{H}}^{\alpha_i}$
		\State Classify:  using 5NN, 5-fold cross-validation, average over 20 runs to yield $\mathbf{AC}_{\alpha_i}$ of every $\alpha_{i}$
		\EndFor
		\EndFor
		\State Choose accuracy:  max-accuracy = $\max\{\mathbf{AC}_{\alpha_i}\}$
	\end{algorithmic}
	\label{al1}
\end{algorithm}
\subsection{The Adaptive Learning Strategy}
\begin{figure*}[htb] 
	\centering
	\includegraphics[scale=0.15]{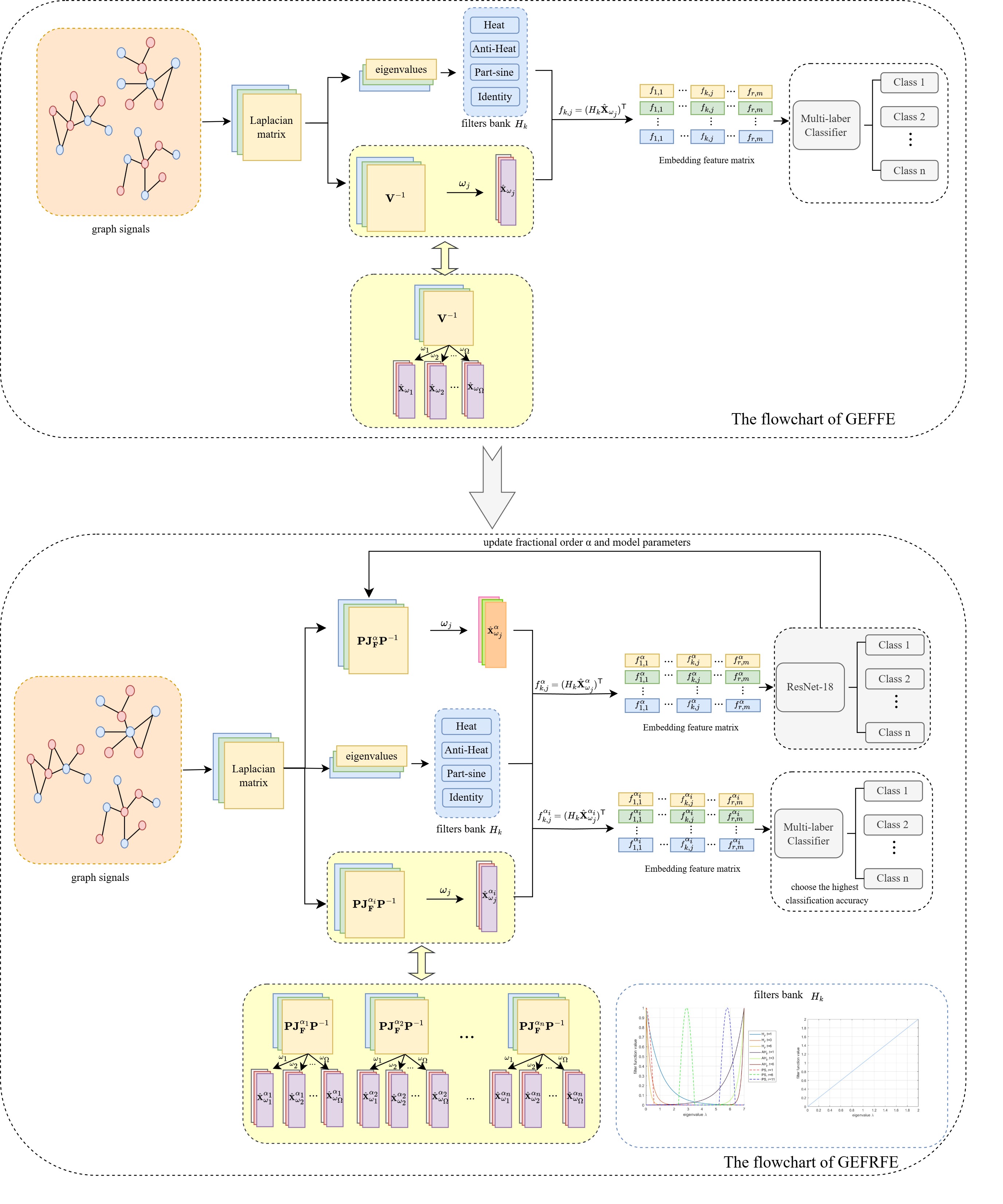}
	\caption{The flowchart of embedding. The upper part of the figure illustrates the existing GEFFE method, whose main limitation lies in performing embeddings solely within the graph spectral domain. This results in a singular embedding space and insufficient exploration of structural information. The lower part of the figure depicts the proposed GEFRFE method, which extends the embedding space from the spectral domain to the graph fractional domain. This addresses the limitation of embedding space singularity and holds the potential to uncover more latent structural information in the fractional domain. }
	\label{flowchart}
\end{figure*}

In addition to the grid search strategy, we also proposed a strategy for adaptively learning the fractional order $\alpha$ of the graph through ResNet-18. This method is based on the existing research \cite{10458263}, which mathematically demonstrates that during the network training stage, the fractional order $\alpha$ of the graph can be updated through backpropagation in any neural network architecture. The detailed steps of the adaptive learning strategy are as follows, as shown in the algorithm \textbf{Algorithm \ref{al2}}. In the classification task, the model is trained using the Adam optimizer and the cross-entropy loss function. The training is conducted for 150 epochs, and an early stopping mechanism (patience = 50) is introduced to prevent overfitting. Additionally, the ReduceLROnPlateau learning rate scheduler is adopted. If the validation loss does not improve for 5 consecutive epochs, the learning rate is multiplied by 0.7. The initial value of the fractional order $\alpha$ is randomly selected from the range $(-3, 3)$ to ensure consistency with the interval used in the grid search. Fig. \ref{flowchart} shows the flowchart of the two methods for updating the fractional order $\alpha$ of the graph. During the experiment, due to the limited number of samples in some datasets, we set the validation set to be consistent with the test set.

All experiments are conducted using PyTorch 1.11 with CUDA 11.3  and the experiment environment is Intel(R) Core(TM) i5-12400F CPU and NVIDIA RTX 4060Ti GPU. 

\begin{algorithm}[htb]
	\caption{Adaptive learning fractional order $\alpha$ by the ResNet-18 method}
	\begin{algorithmic}[1]
		\renewcommand{\algorithmicrequire}{\textbf{Input:}}
		\renewcommand{\algorithmicensure}{\textbf{Output:}}
		\Require Adjacency matrices $\mathbf{A}$, labels, initial fractional order $\alpha $, some power orders $\omega$, some filters $H_k$
		\Ensure Classification accuracy
		
		\State  Compute the Laplacian matrices $\mathbf{L}$: $\mathbf{L} = \mathbf{D} - \mathbf{A}$
		
		\State ResNet-18: compute   $\mathbf{L} = \mathbf{V} \mathbf{J}_{\mathbf{L}} \mathbf{V}^{-1}$,
		$\mathbf{F}^{\alpha} = \mathbf{P} \mathbf{J}^{\alpha}_{\mathbf{F}} \mathbf{P}^{-1}$ and 
		$\hat{\mathbf{x}}^{\alpha} = \mathbf{F}^{\alpha} \mathbf{x}$
		\For{each power order $\omega_{j}$}
		\State $\mathbf{f}^{\alpha}_{k,\omega_j} = H_k[\lambda_l] \hat{\mathbf{X}}^{\alpha}_{\omega_j}[\lambda_l]$,  and $\mathcal{E}_{\mathcal{H}}^{\alpha}$
		\EndFor
		\State Calculate loss: the cross-entropy loss function
		\State Back-propagation: update fractional order $\alpha$ and model weight
		\State Evaluate the model and apply early stopping if necessary 
		\State Load the optimal model parameters
		\State Output the classification accuracy
	\end{algorithmic}
	\label{al2}
\end{algorithm} 
\section{EXPERIMENTAL RESULTS} \label{EXPERIMENTAL}
In this section, we present some experimental results to demonstrate the effectiveness of GEFRFE in various graph classification tasks. The experiments in this paper utilized a diverse set of datasets collected from relevant papers. The attributes of these datasets are shown in Table \ref{tab1}\footnote{\url{https://chrsmrrs.github.io/datasets/docs/datasets/}}.
\begin{itemize}
	\item \textbf{PROTEINS}: This is a representative bioinformatics dataset where graph structures represent protein molecules. An edge exists between two amino acid nodes if the distance between them is less than 6 \AA. The classification task is to categorize proteins as either enzymes or non-enzymes \cite{borgwardt2005protein}.
	\item \textbf{MUTAG}: This dataset consists of nitroaromatic compounds, where nodes represent atom types and edges denote chemical bonds between atoms. The experimental task is a binary classification based on the mutagenicity of the compounds on \textit{Salmonella typhimurium} \cite{debnath1991structure}.
	\item \textbf{PTC-MR}: Standing for the Predictive Toxicology Challenge, this subset (MR) contains chemical molecular structures tested for toxicity in male rats. The objective is to predict the carcinogenicity of these molecules \cite{li2012effective}.
	\item \textbf{IMDB-B (IMDB-BINARY)}: This is a social network dataset derived from collaboration graphs of movie actors. Nodes represent actors, and an edge indicates that two actors co-appeared in the same movie. The dataset describes collaboration networks across two genres: Action and Romance \cite{yanardag2015deep}.
	\item \textbf{IMDB-M (IMDB-MULTI)}: As an extended version of IMDB-B, this dataset further covers three movie genres (e.g., Comedy, Drama, and Sci-Fi), serving as a challenging multi-class classification task \cite{yanardag2015deep}.
\end{itemize}

\begin{table*}[htb]
	% \captionsetup{labelfont={color=blue}}
	\centering
	\caption{Real Graph Datasets}
	\label{tab1}
	\begin{tabular*}{0.8\textwidth}{@{\extracolsep{\fill}} lccccc @{}}
		\toprule
		Datasets & PROTEINS & MUTAG & PTC-MR & IMDB-B & IMDB-M \\
		\midrule
		Graphs & 1113 & 188 & 344 & 1000 & 1500 \\
		Classes & 2 & 2 & 2 & 2 & 3 \\
		Nodes Max & 620 & 28 & 109 & 136 & 89 \\
		Nodes Avg & 39.06 & 17.93 & 14.29 & 19.77 & 13.00 \\
		Edges Avg & 72.82 & 19.79 & 14.69 & 13.06 & 65.93 \\
		\bottomrule
	\end{tabular*}
\end{table*}   

\subsection{The Proposed Filter Function Sets}
\begin{figure}[ht] 
	\subfloat[]{
		\centering
		\includegraphics[height=3.8cm,width=5cm]{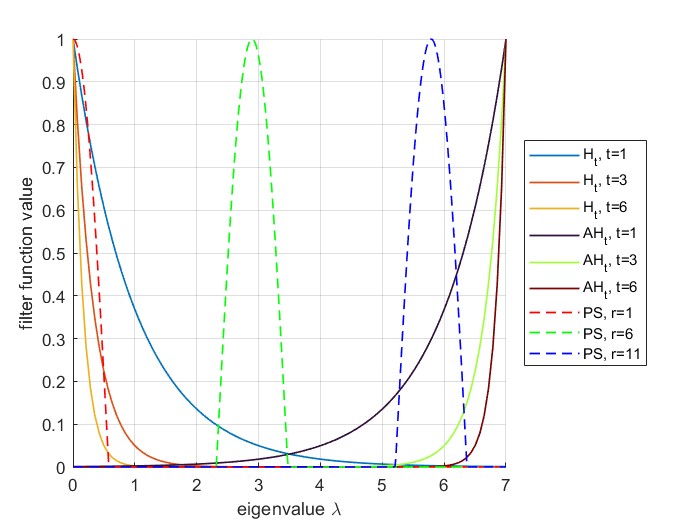}
	}
	\subfloat[]{
		\centering
		\includegraphics[height=3.8cm,width=3.8cm]{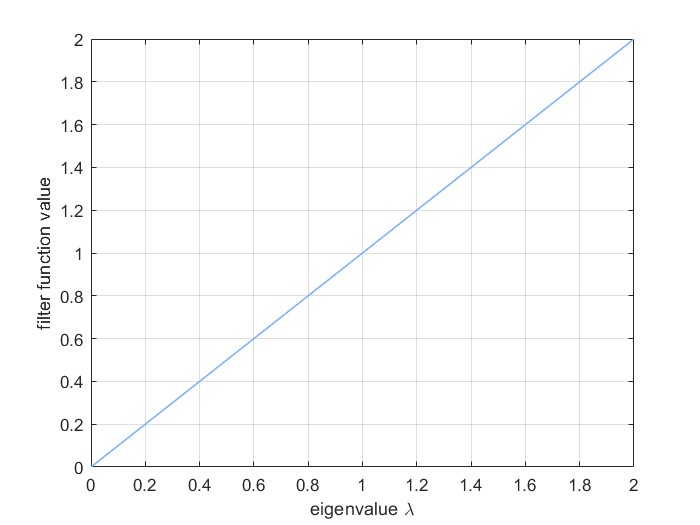}
	}
	\caption{The proposed filter function sets. (a): H, AH, PS filters. (b): X filter.}
	\label{filters}
\end{figure}

In this part, we set some filter functions following the Eq. \eqref{eq8}. The first filter function can be expressed as:
\begin{equation}
	{H}_{t}[\lambda_{l}] = e^{-\lambda_{l}t},
\end{equation} where $t \in \{1,3,6\}$. These filters amplify the low frequencies and decay the high frequencies. The anti-heat filter can be written as:
\begin{equation}
	{AH}_{t}[\lambda_{l}] = e^{-(R-\lambda_{l})t},
\end{equation} where $R$ is the maximum eigenvalues. In these filters, the emphasis is on the high frequencies. The third filter is the part-sine filter set whose every member emphasize on a special portion of the spectrum and can be written as:
\begin{equation}
	PS_{r,\rho} = 
	\begin{cases}
		\sin{\frac{\pi}{2\rho}}(\lambda_{l} - \rho(r -2)) & \text{if}~    \rho (r-2) \leq \lambda_{l} \leq \rho r \\
		0 & \text{otherwise},
	\end{cases}
\end{equation} where $\rho$ is the number of sub-ranges the part-sine functions defined on and $r$ is their sequence number \cite{bahonar2019graph}. When the
most effective frequencies are unknown or the important information spread over the entire frequency domain, the part-sine filters are good candidates.
The last filter is $X(\lambda_{l}) = \lambda_{l}$. Four kinds of filters are shown in Fig. \ref{filters}.

\subsection{The Comparison Between Different Filters } \label{parta}
\begin{figure}[ht] 
	\centering
	\hspace*{-0.5cm}  % 左移 1cm，可根据需要调整
	\includegraphics[height=10cm,width=10cm]{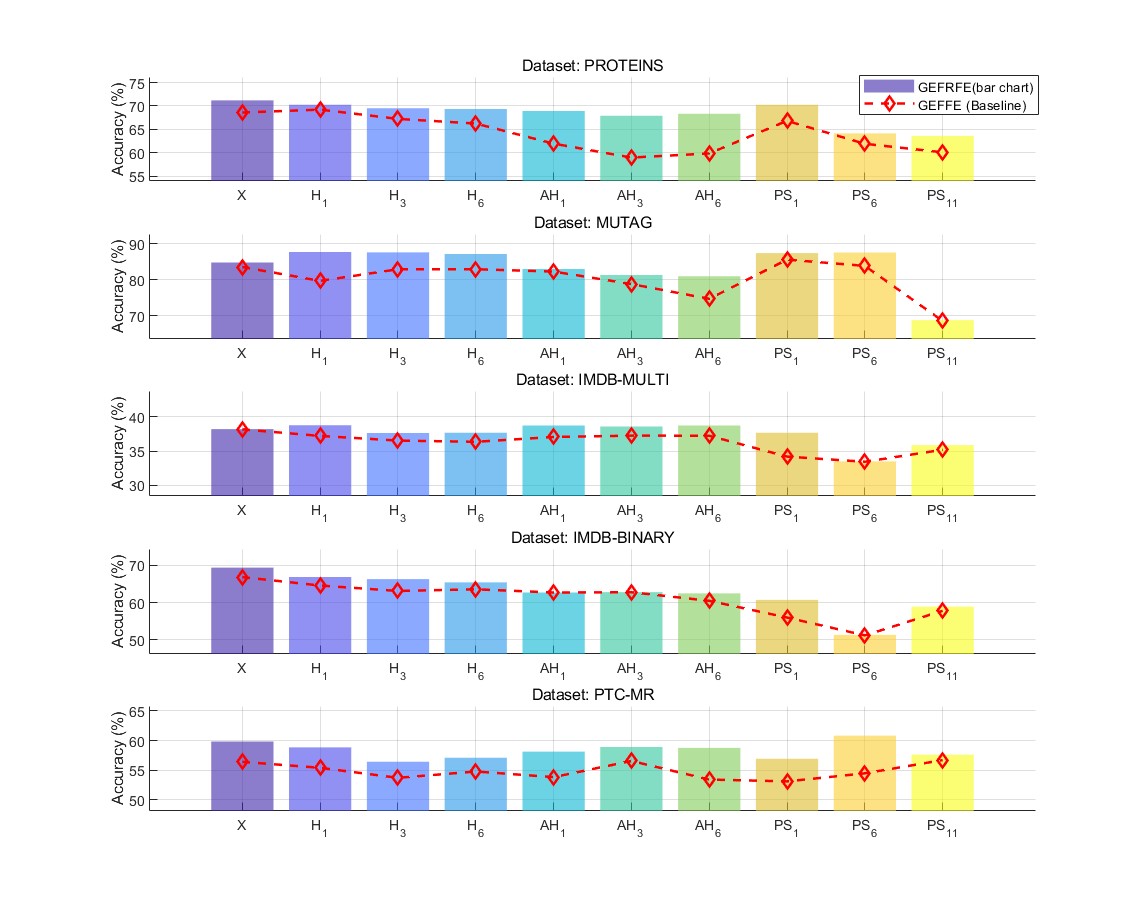}
	%	\subfloat[]{
		%		\centering
		%		\includegraphics[height=3.5cm,width=4.2cm]{Fig3_GEFFE.jpg}
		%	}
	\vspace{-25pt}
	\caption{Comparison of classification performance between GEFRFE and GEFFE.
		The bar chart (GEFRFE) illustrates the maximum classification accuracy of different filters under their respective optimal fractional orders $\alpha$, where the optimal $\alpha$ is determined independently for each filter on a given dataset. The line plot (GEFFE) provides a baseline by showing the accuracy of the same filters at the standard fractional  $\alpha = 1$  }
	\label{fig_2}
\end{figure}

The purpose of this experiment is to investigate how the introduction of the fractional order $\alpha$ contributes to improved classification accuracy, and to demonstrate that different filters capture distinct structural characteristics of the graph.The GEFFE is choosen as baseline method. Fig. \ref{fig_2} illustrates classification results obtained using individual filter from the set $\{X, H_{1}, H_{3}, H_{6}, AH_{1}, AH_{3}, AH_{6}, PS_{1,0.579}, PS_{6,0.579}$$,\\ PS_{11,0.579}\}$ with $\omega = 4$. Experimental results reveal that the GEFRFE consistently outperforms or matches GEFFE across all tested datasets and filter configurations. This validates that performing embedding in the graph fractional domain allows for a more flexible capture of latent local features within graph signals, thereby enhancing the discriminative power of the learned representations.

Specifically, the performance gain of GEFRFE is most pronounced on the MUTAG dataset. Notably, under the $H_{1}$ filter, the GEFRFE achieves a classification accuracy of $87.70\%$, marking an improvement of approximately $7.98\%$ over the $79.72\%$ achieved by GEFFE. This significant enhancement suggests that for molecular graphs with specific topological structures (such as MUTAG), the fractional domain provides a more refined feature characterization than the traditional spectral domain. Furthermore, with the $PS_6$ filter, the GEFRFE maintains a high accuracy of $87.59\%$, significantly exceeding the baseline GEFFE's $83.87\%$. On the IMDB-BINARY dataset, using the $X$ filter, the accuracy of GEFRFE reaches $69.37\%$, a $2.54\%$ increase over GEFFE. On the PROTEINS dataset, the GEFRFE's accuracy exceeds $71\%$ across multiple filters such as $X$ and $PS1$, demonstrating a clear competitive advantage over GFT, which typically hovers around $68\%$.

On the other hand, the five datasets exhibit varying performance across different filters. In the PROTEINS dataset, classification performance fluctuates significantly with changes in the filter. The best performance is observed with the $X$ filter, reaching $71.18\%$, while the worst is recorded with the $PS_{11}$ filter at $63.61\%$, resulting in a performance gap of $7.57\%$. This substantial variance indicates that for biological macromolecule graphs with complex 3D structures, filter $X$ is more effective at preserving global topological features, whereas filter $PS_{11}$ leads to the loss of critical structural information. The sensitivity of the MUTAG dataset to filter selection is even more pronounced. This dataset achieves its peak accuracy of $87.70\%$ with the $H_1$ filter, yet the accuracy drops sharply to $68.83\%$ under the $PS_{11}$ filter, representing a disparity of $18.87\%$. Such considerable performance differences clearly illustrate that the choice of filter profoundly impacts the retention of graph structural features, which ultimately dictates the final classification results.

The above experiments indicate that these independent filters have an impact on the specific structural information of the graph. Therefore, the next step is to explore the influence of filter combinations on the classification results.

\begin{figure}[!h]
	\centering
	\includegraphics[width=3.4in]{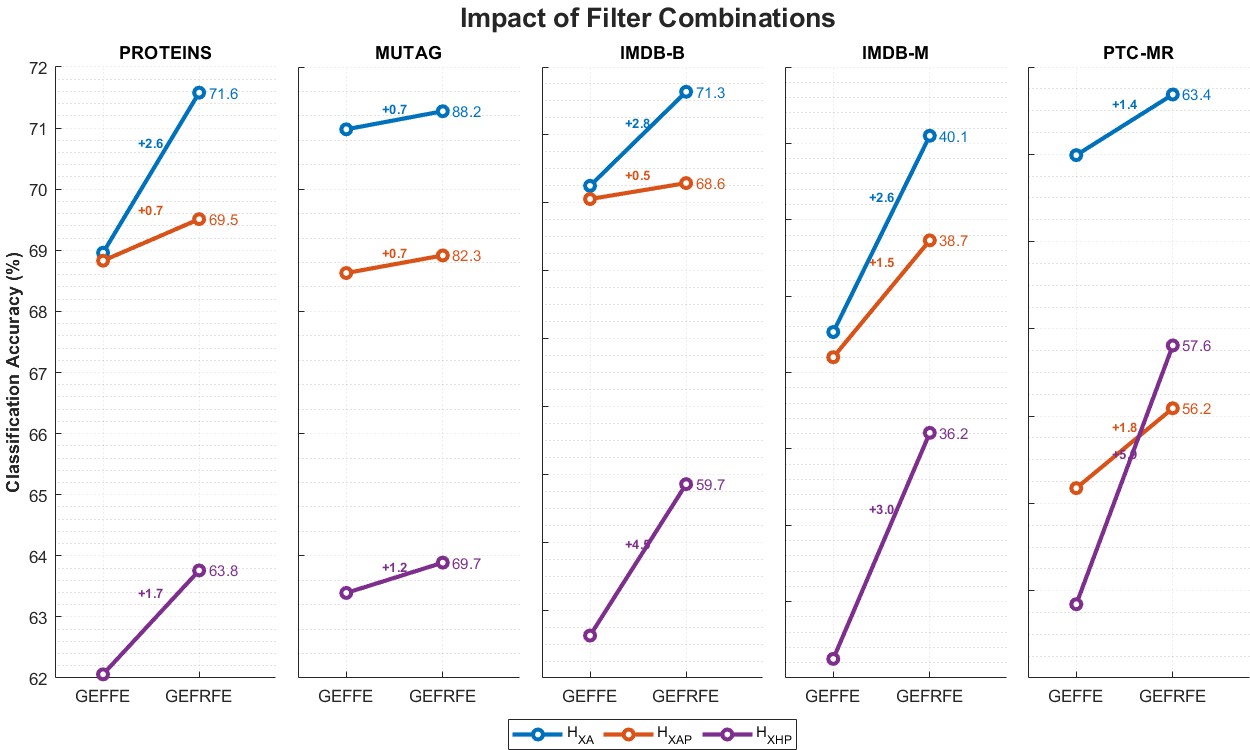}
	\caption{Filter combinations with three different combination strategies.}
	\label{fig_3}
\end{figure}

With a fixed power order $\omega = 3$, Fig. \ref{fig_3} plots the classification accuracy rates of three different filter combinations. The specific filter combinations are as follows: $H_{XA} = X \cup \{AH_{1}, AH_{3}, AH_{6}\}$, $H_{XAP} = X \cup \{AH_{1}, AH_{3}, AH_{6}\} \cup \{PS_{1}, PS_{6}, PS_{11}\}$, and $H_{XHP} = X \cup \{H_{1}, H_{3}, H_{6}\} \cup \{PS_{1}, PS_{6}, PS_{11}\}$. We emphasize the experimental process through the following steps:
\begin{enumerate}
	[label=Step \arabic*:, left=0.5em, labelsep=1em, itemsep=0.5em, font=\normalfont] 
	\item  For each filter, the fractional order $\alpha$ is traversed within the range of $(-3,3)$ with a step size of 0.02, with each fractional order $\alpha$ corresponding to a classification accuracy. The feature with the highest classification accuracy is selected and saved.
	\item  Combine the feature complying with Eq. \eqref{eq13}.
\end{enumerate}

These steps ensure that the fractional order $\alpha$ is within the range of $(-3, 3)$, enabling us to obtain the optimal embedded features. Saving the best features after embedding and then combining them to complete the classification task is quite different from directly concatenating the filters and then performing feature embedding for classification. The latter cannot guarantee that the features selected by each filter when $\alpha$ takes values in $(-3, 3)$ are the optimal ones. However, for GEFFE, both of these methods are feasible because the fractional order $\alpha = 1$ does not change. As shown in Fig. \ref{fig_3}, the $H_{XAP}$ filter combination performs the worst among the five datasets, and the final classification accuracy is not as good as that of $H_{XA}$. Moreover, the classification accuracy of the filter combination $H_{XAP}$ is better than that of the filter combination $H_{XHP}$. This indicates that it is not the case that the more feature concatenation, the better. Different features also have mutual promoting or inhibitory effects. Therefore, choosing the appropriate features for combination is very important.
\subsection{Comparisons among Different Power Orders $\omega$  }\label{partb}
\begin{figure*}[h] 
	\centering
	\subfloat[]{
		\centering
		\includegraphics[height=3.5cm,width=3.5cm]{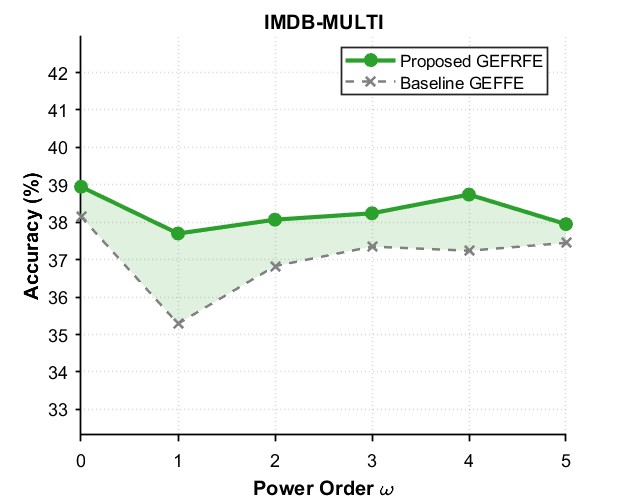}
	}
	%	\hspace{0.5cm}
	\subfloat[]{
		\centering
		\includegraphics[height=3.5cm,width=3.5cm]{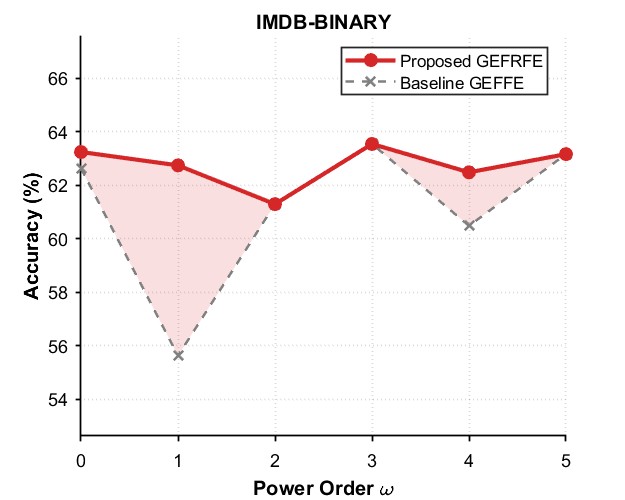}
	}
	%	\hspace{0.5cm}
	%    \\
	\subfloat[]{
		\centering
		\includegraphics[height=3.5cm,width=3.5cm]{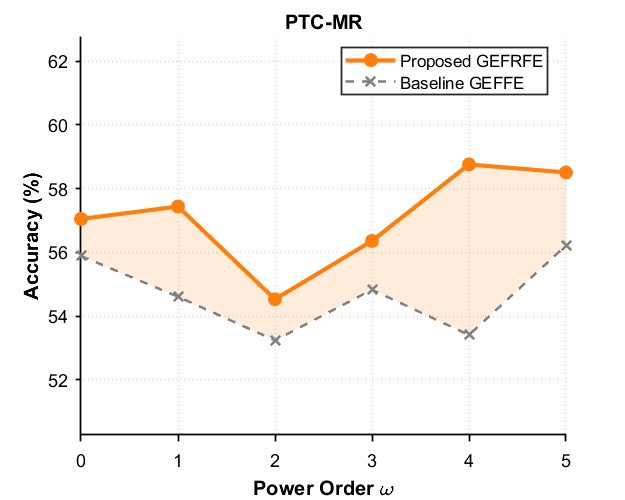}
	}
	%	\\
	\subfloat[]{
		\centering
		\includegraphics[height=3.5cm,width=3.5cm]{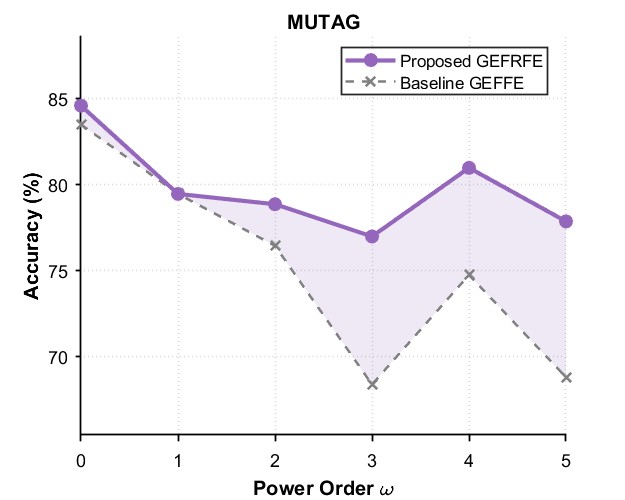}
	}
	%	\hspace{0.5cm}
	\subfloat[]{
		\centering
		\includegraphics[height=3.5cm,width=3.5cm]{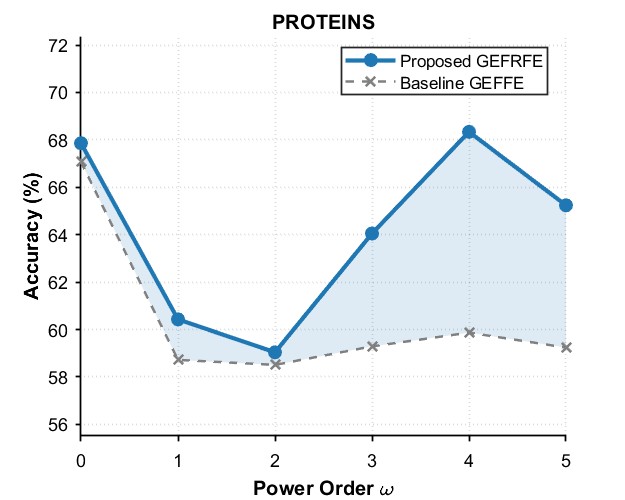}
	}
	\centering
	\caption{The classification accuracies of different power order $\omega$.}
	\label{fig_4}
\end{figure*}

The main purpose of this experiment is to verify that when using GEFRFE, changing the power order $\omega$ still helps to uncover the hidden structural information of the graph. This is supported by proving that different exponents $\omega$ have different classification accuracy rates. By fixing the filter $AH_6$ and using one of the power order $\omega \in \{0, 1, 2, 3, 4, 5\}$ each time. Figs. \ref{fig_4}(a)--(e) respectively show the classification accuracy rates of five datasets using the GEFRFE and GEFFE methods. The shaded areas in each figure represent the gap between the two methods. On all tested datasets, the classification accuracy curve of GEFRFE is always above the baseline method GEFFE curve. This indicates that in the graph frequency domain, the graph embedding is completed, and the model can obtain better graph signal structure features than in the traditional graph domain. Especially on the PROTEINS dataset, when $\omega=4$, the GEFRFE achieves a precision of $68.34\%$, while GEFFE is only $59.87\%$, achieving a significant improvement of up to $8.47\%$. Additionally, the power order $\omega$ plays a role in regulating feature extraction: the experimental results show that the classification accuracy does not change linearly with $\omega$, but presents a clear nonlinear trend. On the MUTAG dataset, the performance reaches its peak at $\omega=0$ ($84.61\%$), then fluctuates, but rebounds to $80.99\%$ at $\omega=4$. On the IMDB-MULTI and PTC-MR datasets, the optimal performance is all around $\omega=4$. This phenomenon indicates that the exponent order $\omega$ effectively mines the discriminative information hidden in the graph structure.
\subsection{The Classification Results of GEFRFE }
\label{partc}
Based on Parts \ref{parta} and \ref{partb}, it can be concluded that different filters and different power orders $\omega$ will result in different classification accuracies, and different combinations of filters will produce different effects. Therefore, it is reasonable for us to combine all possible embedded features to fully explore the hidden structure of the graph and achieve the highest classification accuracy. 

Table \ref{tab2} lists some of the optimal combinations of filters and power $\omega$ selected through the forward selection method from all candidate combinations.
\begin{table*}[htbp]
	\centering % 代替 center 环境，减少多余间距
	\begin{threeparttable}
		\caption{The combination of filters and power order $\omega$}
		\label{tab2}
		\renewcommand{\arraystretch}{1.3} % 稍微收紧行高，更美观
		\begin{tabularx}{\textwidth}{l >{\raggedright\arraybackslash}X >{\raggedright\arraybackslash}X}
			\toprule
			Method & GEFFE-forward & GEFRFE-forward \\ 
			\midrule
			%			PROTEINS & X-2, X-3, H1-4, PS6-4 & X-4, H6-5, AH6-3 \\ 
			PROTEINS & $X-2$, $X-3$, $H_1-4$, $PS_6-4$ & $X-4$, $H_6-5$, $AH_6-3$ \\
			\addlinespace % 增加数据集之间的微小间距
			IMDB-MULTI & $AH_6-0$, $H3-1$, $AH1-5$, $AH_1-2$, $PS_6-0$, $AH_6-3$, $AH_1-3$, $H_3-3$, $H_3-4$, $AH_3-4$ & $AH_6-4$, $PS_11-1$, $PS_1-4$, $PS_11-5$ \\ 
			\addlinespace
			IMDB-BINARY & $X-0$, $AH_1-5$ & $X-2$, $PS_6-0$, $PS_{11}-5$, $PS_6-4$, $H_1-4$, $H_6-1$ \\ 
			\addlinespace
			PTC-MR & $AH_3-5$, $PS_{11}-3$, $X-1$ & $PS_6-4$, $PS_{11}-4$ \\ 
			\addlinespace
			MUTAG & $PS_1-4$, $AH_3-3$, $AH_3-2$, $H_3-5$, $AH_6-3$, $PS_{11}-3$, $PS_{11}-2$ & $H_1-4$, $PS_{11}-1$, $PS_{11}-0$ \\ 
			\bottomrule
		\end{tabularx}
		\begin{tablenotes}
			\small % 注释字体调小，符合规范
			\item[*] $PS_{i,0.579}$ is abbreviated as $PS_{i}$ in the table.
			\item[*] The suffix after the hyphen (e.g., $5$ in $AH_3-5$) represents the power order $\omega$.
		\end{tablenotes}
	\end{threeparttable}
\end{table*}
The classification accuracy corresponding to the forward selection method is listed in Table \ref{tab:4}, In GEFFE-all and GEFRFE-all, they respectively represent all the embedded features that are combined with the power order $\omega$ using the filters of GEFFE and GEFRFE. GEFFE-forward and GEFRFE-forward respectively represent the highest accuracy rates achieved during the forward selection process using GEFFE and GEFRFE. Specifically:
\begin{enumerate}
	[label=Step \arabic*:, left=0.5em, labelsep=1em, itemsep=0.5em, font=\normalfont] 
	\item The embedded feature that yields the highest classification accuracy is selected first. 
	\item At each step, the feature that brings the greatest improvement in accuracy among the remaining ones is added, until no further increase in accuracy can be achieved.
\end{enumerate}

Additionally, we adopted the ResNet-18 for adaptive learning of the fractional order $\alpha$. \textbf{Algorithm \ref{al2}} provides the detailed steps. Table \ref{tab3} lists the model parameters corresponding to each dataset.
\begin{table*}[htp]
	\begin{center}
		\caption{ResNet-18 parameters setting}
		\renewcommand{\arraystretch}{1.2}
		\label{tab3}
		\begin{threeparttable}
			\begin{tabular}{ c  c  c  c c}
				\hline
				Datasets  &  learning rate (network) & learning rate ($\alpha$) & weight decay  & batch size  \\ 
				\hline			
				PROTEINS & $1 \times 10^{-5}$ & $1 \times 10^{-3}$ & $1 \times 10^{-3}$ & $32$ \\ 
				IMDB-MULTI & $ 1 \times 10^{-5}$  & $1 \times 10^{-3}$ & $1 \times 10^{-3}$ & $32$\\
				IMDB-BINARY &  $ 5 \times 10^{-5}$  & $1 \times 10^{-3}$ & $1 \times 10^{-3}$ & $16$\\
				PTC-MR &  $ 1 \times 10^{-5}$  & $1 \times 10^{-3}$ & $1 \times 10^{-3}$ & $16$\\
				MUTAG &  $ 1 \times 10^{-5}$  & $1 \times 10^{-3}$ & $1 \times 10^{-3}$ & $16$\\
				\hline
			\end{tabular}
			%			\begin{tablenotes} 
				%				\item --- Represents no weight decay
				%			\end{tablenotes}
		\end{threeparttable}
	\end{center}
\end{table*}

To avoid potential misunderstandings, we conducted an additional experiment in which the initial value of the graph fractional order is set to $\alpha = 1$ and is not involved in the network training process. This experiment aims to prove that the improvement in classification accuracy is mainly due to the introduction of the graph fractional order $\alpha$, rather than replacing the 5NN with the ResNet-18 model. It is particularly important to note that in this strategy, the filters are first concatenated and then the graph feature embedding is performed before completing the classification. This was an inevitable choice because the graph fractional order $\alpha$ is continuously optimized throughout the experiment and could not preserve the optimal results corresponding to different filters. 

\begin{figure}[ht] 
	\centering
	\includegraphics[height=8cm,width=8cm]{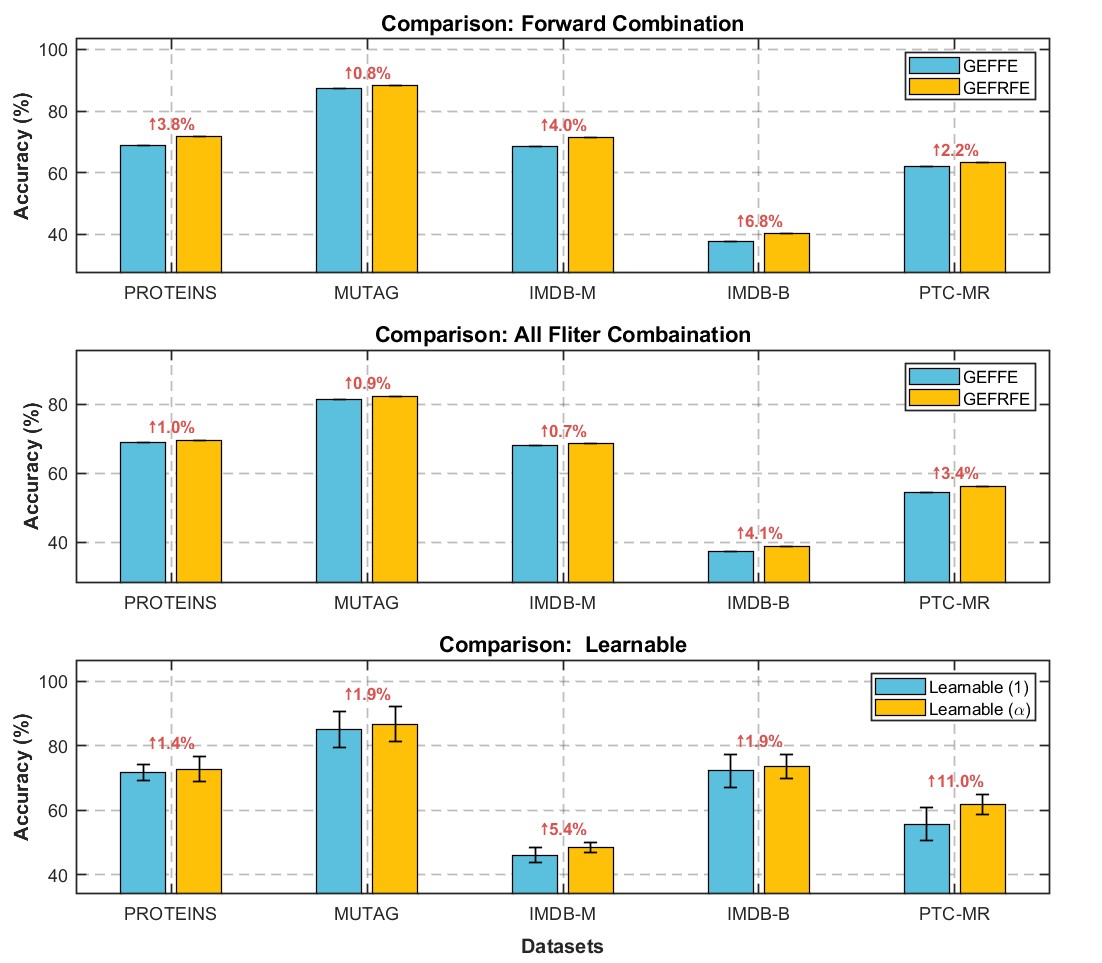}
	%	\subfloat[]{
		%		\centering
		%		\includegraphics[height=3cm,width=4.2cm]{forward1.jpg}
		%	}
	\caption{The classification accuracy of adaptive learning, forward selection, and the combination of all filters.}
	\label{fig_6}
\end{figure}

Fig. \ref{fig_6} presents a bar chart showing the differences in classification accuracy caused by the fractional order $\alpha$. From the figure, it can be clearly seen that the classification accuracy of GEFRFE-learnable($\alpha$) is higher than that of GEFRFE-learnable($1$), directly indicating that introducing the graph fractional order $\alpha$ can retain more graph structure information of the embedding, thereby achieving a higher classification accuracy.
\begin{table*}[htp]
	\begin{center}
		\caption{All results of six methods}
		\renewcommand{\arraystretch}{1.2} %
		\label{tab:4}
		\begin{threeparttable}
			\begin{tabular}{ c  c  c  c c c c}
				\hline
				\tabcolsep=0.2cm
				Method  &  PROTEINS  & MUTAG  & PTC-MR & IMDB-BINARY & IMDB-MULTI 
				\tabcolsep=0.2cm
				\\ 
				\hline
				GEFFE-all & 68.83\% & 81.57\% & 54.35\% & 68.11\% & 37.20\%  \\ 
				GEFFE-forward& 68.96\% & 87.45\% & 61.98\% & 68.50\% & 37.53\%   \\ 
				GEFRFE-all & 69.51\% & 82.29\% & 56.18\% & 68.58\% & 38.73\%   \\ 
				GEFRFE-forward & $\bm{71.58\%}$ & $\bm{88.19\%}$ & $\bm{63.37\%}$ & $\bm{71.27\%}$ & $\bm{40.10\%}$ \\
				%				GEFFE-learnable($1$) & $71.75 \pm 2.42$ & $85.15 \pm 5.59$ & $\55.77 \pm 5.11$ & $72.20 \pm 5.25$ & $46.13 \pm 2.36$ \\
				%				GEFRFE-learnable($\alpha$) & $\mathbf{72.79 \pm 3.93}$ & $\mathbf{86.73 \pm 5.34}$ & $\mathbf{61.90 \pm 3.11}$ & $\mathbf{73.60 \pm 3.69}$ & $\mathbf{48.60 \pm 1.59}$ \\
				\hline
			\end{tabular}
			%			\begin{tablenotes} 
				%				\raggedright % Align the footnote text to the left
				%%				\item 最高分类精度使用加粗表示
				%			\end{tablenotes}
		\end{threeparttable}
	\end{center}
\end{table*}

Finally, we compare our method with the classic and current mainstream graph embedding approaches, aiming to explore whether introducing the GFRFT into graph embedding is a correct and meaningful direction. This included two graph kernel methods: GK \cite{shervashidze2009efficient} and WL kernel \cite{shervashidze2011weisfeiler}. Two methods based on GNN: DGCN \cite{zhang2018end} and GCN \cite{kipf2016semi}. And the current mainstream graphformer methods: Transformer \cite{vaswani2017attention}, Graphormer \cite{ying2021transformers}, and SGFormer \cite{wu2023sgformer}. In the experiments, these methods used the same set of data partitions and 10-fold cross-validation, and the final results were averaged.

The comparative experiments on five public graph classification datasets show, as presented in Table \ref{tab:experimental_results}, that the proposed GEFRFE-learnable($\alpha$) method achieves comparable accuracy levels with mainstream GNN and Graph Transformer methods. At the same time, it demonstrates unique theoretical and practical value. On datasets such as  MUTAG, PTC-MR and IMDB-BINARY, the accuracy of GEFRFE-learnable($\alpha$) ($86.73\%$, $61.90\%$, $74.60\%$) consistently ranks within the top one-third of the comparison methods, significantly outperforming early kernel methods (such as GK) and basic Transformer baselines (such as Transformer, Graphormer), and being comparable to baselines such as SGFormer and GCN. This result indicates that integrating GFRFT into the graph embedding field is a correct and effective research path.
%\begin{table*}[htp]
%	\centering
%	\caption{Comparison of accuracy rates of different methods on five image classification datasets (\%)}
%	\label{tab:experimental_results}
%	\small
%	\renewcommand{\arraystretch}{1.2}
%	\begin{tabularx}{\textwidth}{l XXXXX}
	%		\toprule
	%		方法 & PROTEINS & MUTAG & PTC-MR & IMDB-BINARY & IMDB-MULTI \\
	%		\midrule
	%		\multicolumn{6}{c}{\textbf{Kernel methods}} \\
	%		GK          & $71.4 \pm 0.3$ & $81.7 \pm 2.1$ & $55.3 \pm 1.4$ & $65.9 \pm 1.0$ & $43.9 \pm 0.4$ \\
	%		WL kernel   & $75.0 \pm 3.1$ & $90.4 \pm 5.7$ & $59.9 \pm 4.3$ & $73.8 \pm 3.9$ & $50.9 \pm 3.8$ \\
	%		\midrule
	%		\multicolumn{6}{c}{\textbf{GNN methods}} \\
	%		DGCNN       & $75.5 \pm 0.9$ & $85.8 \pm 1.8$ & $58.6 \pm 2.5$ & $70.0 \pm 10.9$ & $47.8 \pm 10.9$ \\
	%		GCN         & $75.2 \pm 2.8$ & $85.1 \pm 5.8$ & $63.1 \pm 4.3$ & $73.8 \pm 3.4$ & $55.2 \pm 0.3$ \\
	%		\midrule
	%		\multicolumn{6}{c}{\textbf{Graph Transformers}} \\
	%		Transformer & $66.3 \pm 8.4$ & $81.9 \pm 9.7$ & $57.3 \pm 7.0$ & $71.1 \pm 3.8$ & $45.8 \pm 3.8$ \\
	%		Graphormer  & $68.5 \pm 2.3$ & $82.5 \pm 3.8$ & $59.2 \pm 4.6$ & $73.5 \pm 3.8$ & $48.9 \pm 2.3$ \\
	%		SGFormer    & $74.6 \pm 3.0$ & $88.6 \pm 6.3$ & $65.2 \pm 4.2$ & $74.7 \pm 4.1$ & $56.4 \pm 3.4$ \\
	%		\midrule
	%		\multicolumn{6}{c}{\textbf{GEFRFE (GEFRFE-learnable($\alpha$))}} \\
	%		GEFFE-learnable($1$) & $71.75 \pm 2.42$ & $85.15 \pm 5.59$ & $55.77 \pm 5.11$ & $72.20 \pm 5.25$ & $46.13 \pm 2.36$ \\
	%		GEFRFE-learnable($\alpha$) & $72.79 \pm 3.93$ & $86.73 \pm 5.34$ & $61.90 \pm 3.11$ & $74.0 \pm 3.69$ & $48.60 \pm 1.59$ \\
	%		\bottomrule
	%	\end{tabularx}
%\end{table*}

\begin{table*}[htp]
	\centering
	\caption{Comparison of accuracy  (Top 3 marked: * for 1st, $\dagger$ for 2nd, $\ddagger$ for 3rd)}
	\label{tab:experimental_results}
	\small
	\renewcommand{\arraystretch}{1.2}
	\begin{tabularx}{\textwidth}{l XXXXX}
		\toprule
		方法 & PROTEINS & MUTAG & PTC-MR & IMDB-BINARY & IMDB-MULTI \\
		\midrule
		\multicolumn{6}{c}{\textbf{Kernel methods}} \\
		GK          & $71.4 \pm 0.3$ & $81.7 \pm 2.1$ & $55.3 \pm 1.4$ & $65.9 \pm 1.0$ & $43.9 \pm 0.4$ \\
		WL kernel   & $75.0 \pm 3.1^{\ddagger}$ & $90.4 \pm 5.7^{*}$ & $59.9 \pm 4.3$ & $73.8 \pm 3.9^{\ddagger}$ & $50.9 \pm 3.8$ \\
		\midrule
		\multicolumn{6}{c}{\textbf{GNN methods}} \\
		DGCNN       & $75.5 \pm 0.9^{*}$ & $85.8 \pm 1.8$ & $58.6 \pm 2.5$ & $70.0 \pm 10.9$ & $47.8 \pm 10.9$ \\
		GCN         & $75.2 \pm 2.8^{\dagger}$ & $85.1 \pm 5.8$ & $63.1 \pm 4.3^{\dagger}$ & $73.8 \pm 3.4^{\ddagger}$ & $55.2 \pm 0.3^{\dagger}$ \\
		\midrule
		\multicolumn{6}{c}{\textbf{Graph Transformers}} \\
		Transformer & $66.3 \pm 8.4$ & $81.9 \pm 9.7$ & $57.3 \pm 7.0$ & $71.1 \pm 3.8$ & $45.8 \pm 3.8$ \\
		Graphormer  & $68.5 \pm 2.3$ & $82.5 \pm 3.8$ & $59.2 \pm 4.6$ & $73.5 \pm 3.8$ & $48.9 \pm 2.3$ \\
		SGFormer    & $74.6 \pm 3.0$ & $88.6 \pm 6.3^{\dagger}$ & $65.2 \pm 4.2^{*}$ & $74.7 \pm 4.1^{*}$ & $56.4 \pm 3.4^{*}$ \\
		\midrule
		\multicolumn{6}{c}{\textbf{GEFRFE (GEFRFE-learnable($\alpha$))}} \\
		GEFFE-learnable($1$) & $71.75 \pm 2.42$ & $85.15 \pm 5.59$ & $55.77 \pm 5.11$ & $72.20 \pm 5.25$ & $46.13 \pm 2.36$ \\
		GEFRFE-learnable($\alpha$) & $72.79 \pm 3.93$ & $86.73 \pm 5.34^{\ddagger}$ & $61.90 \pm 3.11^{\ddagger}$ & $74.60 \pm 3.69^{\dagger}$ & $51.60 \pm 1.59^{\ddagger}$ \\
		\bottomrule
	\end{tabularx}
\end{table*}

\subsection{The Comparison of Computational Complexity} \label{Complexity}
In terms of  computation complexity, it should be noted that in graph signal the feature vector length is $N$. The GEFFE involves the eigendecomposition of the Laplacian matrix $\mathbf{L}$, GFT and filtering, their computation complexity are $\mathcal{O}(N^3)$, $\mathcal{O}(N^2)$ and $\mathcal{O}(N)$, respectively. The GEFRFE involves the eigendecomposition of the Laplacian matrix $\mathbf{L}$ and GFT matrix $\mathbf{V}^{-1}$, GFRFT, and filtering, their computation complexity are $\mathcal{O}(2N^3)$, $\mathcal{O}(N^2)$ and $\mathcal{O}(N)$, respectively. Thus, The overall computational complexity of both GEFFE and GEFRFE is $\mathcal{O}(N^3)$. In practical experiments, the GEFRFE and the GEFFE are conducted simultaneously, as the GEFFE can be regarded as a special case of the GEFRFE when $\alpha = 1$. The time difference between the GEFRFE and the GEFFE primarily depends on the number of the fractional order $\alpha$.  For the grid search strategy, selecting a suitable search range and interval can significantly improve efficiency and reduce computational time.
For the approach utilizing the ResNet-18 with adaptive learning of the fractional order $\alpha$, both the number of training epochs and the batch size significantly influence the overall runtime. 

\section{Conclusion and future work}
This study introduces GEFRFE, a novel graph embedding framework operating in the graph fractional domain. The proposed approach integrates graph fractional domain filtering and the power form of Laplacian eigenvector elements, where each filter-power combination yields an individual feature. Multiple features are selected through the forward selection mechanism and combined to form the final embedding representation for completing the classification task. To determine the fractional order in the graph fractional domain, two strategies are proposed: a grid search strategy and an adaptive learning strategy. The grid search allows flexible configuration of search range and interval based on specific tasks, while the adaptive learning strategy incorporates a ResNet-18 network to mitigate potential gradient vanishing issues during training. Empirical results across five benchmark datasets confirm that the GEFRFE captures richer structural features. Additionally, the GEFRFE maintains computational complexity on par with GEFFE, ensuring practical scalability.

Compared with the existing methods, the core advantage of GEFRFE lies in its breakthrough in the representation paradigm: it is the first to expand the graph representation space from the discrete vertex domain and graph spectrum domain ($\alpha=1$) to the continuous graph fractional transformation domain. By adaptively learning the fractional  $\alpha$, it achieves a more comprehensive capture of graph structure information. Although the accuracy has not achieved a comprehensive superiority, this method breaks through the fixed domain limitations of traditional spectral graph embedding, providing a new paradigm for the development of spectral graph embedding from "fixed domain" to "generalized domain". As the first graph embedding framework based on the GFRFT, this work verifies the potential of fractional-domain representation and opens up an important direction for subsequent research.

Future work will focus on further optimizing  the GEFRFE framework. Since the GEFRFE method relies on the spectral
decomposition operations, it faces efficiency bottlenecks when scaling to large graphs. It is the first to expand the representation space from the discrete vertex domain and the graph spectrum domain to the continuous graph fractional transformation domain, breaking through the fixed domain limitations of traditional spectral graph embedding and achieving a more comprehensive capture of graph structure information. Future research could focus on integrating GFRFT or GEFRFE into various feature extraction tasks, which might potentially enhance performance in complex graph representation learning tasks.
\bibliographystyle{IEEEtran} % 指定 IEEE 标准样式
\bibliography{ref}%.bib文档名

@article{ortega2018graph,
	title={Graph signal processing: Overview, challenges, and applications},
	author={Ortega, A. and Frossard, P.  and Kova{\v{c}}evi{\'c}, J.  and Moura, J. M. F. and Vandergheynst, P.},
	journal={Proc. IEEE Proc. IRE},
	volume={106},
	number={5},
	pages={808--828},
	year={2018},
	publisher={IEEE}
}

@article{sandryhaila2014big,
	title={Big data analysis with signal processing on graphs: Representation and processing of massive data sets with irregular structure},
	author={Sandryhaila, A. and Moura, J. M. F.},
	journal={IEEE Signal Process. Mag.},
	volume={31},
	number={5},
	pages={80--90},
	year={2014},
	publisher={IEEE}
}

@article{wang2017knowledge,
	title={Knowledge graph embedding: A survey of approaches and applications},
	author={Wang, Q. and Mao, Z. D. and Wang, B. and Guo, L.},
	journal={IEEE Trans. Knowl. Data Eng.},
	volume={29},
	number={12},
	pages={2724--2743},
	year={2017},

}

@article{li2021learning,
	title={Learning knowledge graph embedding with heterogeneous relation attention networks},
	author={Li, Z. F. and Liu, H. and Zhang, Z. L. and Liu, T. T. and Xiong, N. N.},
	journal={IEEE Trans. Neural Netw. Learn. Syst.},
	volume={33},
	number={8},
	pages={3961--3973},
	year={2021},
	publisher={IEEE}
}

@article{8937811,
	author={Ienco, D. and Pensa, R. G.},
	journal={IEEE Trans. Neural Netw. Learn. Syst.}, 
	title={Enhancing Graph-Based Semisupervised Learning via Knowledge-Aware Data Embedding}, 
	year={2020},
	volume={31},
	number={11},
	pages={5014-5020},
	}

@article{9741755,
	author={Zhang, H. Y. and Li, P. and Zhang, R. and Li, X. L.},
	journal={IEEE Trans. Neural Netw. Learn. Syst.}, 
	title={Embedding Graph Auto-Encoder for Graph Clustering}, 
	year={2023},
	volume={34},
	number={11},
	pages={9352-9362},
	}

@article{deng2019graphzoom,
	title={Graphzoom: A multi-level spectral approach for accurate and scalable graph embedding},
	author={Deng, C. H. and Zhao, Z. Q. and Wang, Y. Y. and Zhang, Z. R. and Feng, Z.},
	journal={arXiv:1910.02370},
	year={2019}
}

@article{luo2003spectral,
	title={Spectral embedding of graphs},
	author={Luo, B. and Wilson, R. C. and Hancock, E. R.},
	journal={Pattern Recognit.},
	volume={36},
	number={10},
	pages={2213--2230},
	year={2003},
	publisher={Elsevier}
}

@article{cai2018comprehensive,
	title={A comprehensive survey of graph embedding: Problems, techniques, and applications},
	author={Cai, H. Y. and Zheng, V. W and Chang, K. C.},
	journal={IEEE Trans. Knowl. Data Eng.},
	volume={30},
	number={9},
	pages={1616--1637},
	year={2018},

}

@article{wu2023survey,
	title={A survey on graph embedding techniques for biomedical data: Methods and applications},
	author={Wu, Y. Z. and Chen, Y. K. and Yin, Z. S. and Ding, W. P. and King, I.},
	journal={ Inf. Fusion },
	volume={100},
	pages={101909},
	year={2023},
	publisher={Elsevier}
}

@article{perozzi2014deepwalk,
	title={Deepwalk: Online learning of social representations},
	author={Perozzi, B. and Al-Rfou, R. and Skiena, S. },
	journal={\textnormal{in} Proc. ACM SIGKDD 20th Int. Conf. Knowl.
	Discov. Data Mining},
	pages={701--710},
	year={2014}
}

@article{chung1997spectral,
	title={Spectral graph theory},
	author={Chung, F. R. and Graham, C. },
	volume={92},
	year={1997},
	journal={A. M. S.}
}

@article{emms2005matrix,
	title={A matrix representation of graphs and its spectrum as a graph invariant},
	author={Emms, D. and Hancock, E. R. and Severini, S. and Wilson, R. C. },
	journal={arXiv preprint quant-ph/0505026},
	year={2005}
}

@article{xiao2009graph,
	title={Graph characteristics from the heat kernel trace},
	author={Xiao, B. and Hancock, E. R. and Wilson, R. C.},
	journal={Pattern Recognit.},
	volume={42},
	number={11},
	pages={2589--2606},
	year={2009},
	publisher={Elsevier}
}

@article{wilson2014graph,
	title={Graph signatures for evaluating network models},
	author={Wilson, R. C.},
	journal={\textnormal{in} ICPR},
	pages={100--105},
	year={2014},
	publisher={IEEE}
}

@article{ren2010graph,
	title={Graph characterization via ihara coefficients},
	author={Ren, P. and Wilson, R. C. and Hancock, E. R.},
	journal={ IEEE Trans. Neural Netw.},
	volume={22},
	number={2},
	pages={233--245},
	year={2010},
	publisher={IEEE}
}

@article{bahonar2019graph,
	title={Graph embedding using frequency filtering},
	author={Bahonar, H. and Mirzaei, A. and Sadri, S. and Wilson, R. C.},
	journal={IEEE Trans. Pattern Anal. Mach. Intell.},
	volume={43},
	number={2},
	pages={473--484},
	year={2019},
	publisher={IEEE}
}

@article{wilson2005pattern,
	title={Pattern vectors from algebraic graph theory},
	author={Wilson, Richard C. and Hancock, Edwin R and Luo, Bin},
	journal={IEEE Trans. Pattern Anal. Mach. Intell.},
	volume={27},
	number={7},
	pages={1112--1124},
	year={2005},
	publisher={IEEE}
}

@article{aziz2013backtrackless,
	title={Backtrackless walks on a graph},
	author={Aziz, F. and Wilson, R. C. and Hancock, E. R.},
	journal={IEEE Trans. Neural Netw. Learn. Syst.},
	volume={24},
	number={6},
	pages={977--989},
	year={2013},
	publisher={IEEE}
}

@article{gartner2003graph,
	title={On graph kernels: Hardness results and efficient alternatives},
	author={G{\"a}rtner, T. and Flach, P. and Wrobel, S.},
	journal={\textnormal{in} Proceedings of COLT-KM},
	pages={129--143},
	year={2003},
	organization={Springer}
}

@article{wang2017fractional,
	title={The fractional {F}ourier transform on graphs},
	author={Wang, Y. Q. and Li, B. Z. and Cheng, Q. Y.},
	journal={\textnormal{in} Proc. Asia-Pacific Signal Inf. Process. Assoc. Annu. Summit
	Conf.},
	pages={105--110},
	year={2017}
}

@article{alikacsifouglu2024graph,
	title={Graph fractional {F}ourier transform: A unified theory},
	author={Alika{\c{s}}ifo{\u{g}}lu, T. and Kartal, B. and Ko{\c{c}}, A.},
	journal={IEEE Trans. Signal Process.},
	year={2024},
	pages={3834--3850},
	publisher={IEEE}
}

@article{wang2018fractional,
	title={The fractional {F}ourier transform on graphs: Sampling and recovery},
	author={Wang, Y. Q. and Li, B. Z.},
	journal={\textnormal{in} Proc. 14th Int. Conf. Signal Process.},
	pages={1103--1108},
	year={2018}
}

@article{wei2024generalized,
	title={Generalized sampling of graph signals with the prior information based on graph fractional Fourier transform},
	author={Wei, D. Y.  and Yan, Z. Y.},
	journal={Signal Process.},
	volume={214},
	pages={109263},
	year={2024},
	publisher={Elsevier}
}

@article{ozturk2021optimal,
	title={Optimal fractional {F}ourier filtering for graph signals},
	author={Ozturk, C. and Ozaktas, H. M. and Gezici, S. and Ko{\c{c}}, A.},
	journal={IEEE Trans. Signal Process.},
	volume={69},
	pages={2902--2912},
	year={2021},
	publisher={IEEE}
}

@article{yan2022multi,
	title={Multi-dimensional graph fractional {F}ourier transform and its application to data compression},
	author={Yan, F. J. and Li, B. Z.},
	journal={Digit. Signal Process.},
	volume={129},
	pages={103683},
	year={2022},
	publisher={Elsevier}
}

@article{zhang2025graph,
	title={The Graph Fractional {F}ourier Transform in Hilbert Space},
	author={Zhang, Y. and Li, B. Z.},
	journal={IEEE Trans. Signal Inf. Process. Netw.},
	year={2025},
	publisher={IEEE}
}

@article{gan2025windowed,
	title={The windowed two-dimensional graph fractional {F}ourier transform},
	author={Gan, Y. C. and Chen, J. Y. and Li, B. Z.},
	journal={Digit. Signal Process.},
	volume={162},
	pages={105191},
	year={2025},
	publisher={Elsevier}
}

@article{yan2021windowed,
	title={Windowed fractional {F}ourier transform on graphs: Properties and fast algorithm},
	author={Yan, F. J. and Li, B. Z.},
	journal={Digit. Signal Process.},
	volume={118},
	pages={103210},
	year={2021},
	publisher={Elsevier}
}

@article{yan2020windowed,
	title={Windowed fractional {F}ourier transform on graphs: Fractional translation operator and Hausdorff-Young inequality},
	author={Yan, F. J. and Gao, W. B. and Li, B. Z. },
	journal={\textnormal{in} Proc. Asia-Pacific Signal Inf. Process. Assoc. Annu. Summit Conf.},
	pages={255--259},
	year={2020},
	organization={IEEE}
}

@article{yan2023spectral,
	title={Spectral graph fractional {F}ourier transform for directed graphs and its application},
	author={Yan, F. J. and Li, B. Z. },
	journal={Signal Process.},
	volume={210},
	pages={pp. 109099},
	year={2023},
	publisher={Elsevier}
}

@article{wei2024hermitian,
	title={Hermitian random walk graph {F}ourier transform for directed graphs and its applications},
	author={Wei, D. Y. and Yuan, S. X.},
	journal={Digit. Signal Process.},
	volume={155},
	pages={104751},
	year={2024},
	publisher={Elsevier}
}

@ARTICLE{6409473,
	author={Sandryhaila, A. and Moura, J. },
	journal={IEEE Trans. Signal Process.}, 
	title={Discrete Signal Processing on Graphs}, 
	year={2013},
	volume={61},
	number={7},
	pages={1644-1656},
	doi={10.1109/TSP.2013.2238935}}

@article{shuman2013emerging,
	title={The emerging field of signal processing on graphs: Extending high-dimensional data analysis to networks and other irregular domains},
	author={Shuman, D. I. and Narang, S. K. and Frossard, P. and Ortega, A. and Vandergheynst, P. },
	journal={IEEE Signal Process. Mag.},
	volume={30},
	number={3},
	pages={83--98},
	year={2013},
	publisher={IEEE}
}

@book{bai2007heat,
	title={Heat kernel analysis on graphs.},
	author={Bai, X.},
	year={2007},
	publisher={PhD thesis, The University
	of York}
}

@article{borgwardt2005protein,
	title={Protein function prediction via graph kernels},
	author={Borgwardt, K. M. and Ong, C. S. and Sch{\"o}nauer, S. and Vishwanathan, S. V. N. and Smola, A. J. and Kriegel, H. -P.},
	journal={Bioinformatics},
	volume={21},
	year={2005},
	publisher={Oxford University Press}
}

@ARTICLE{10458263,
	author={Koç, E. and Alikaşifoğlu, T. and Aras, A. C. and Koç, A.},
	journal={IEEE Signal Process. Lett.}, 
	title={Trainable Fractional Fourier Transform}, 
	year={2024},
	volume={31},
	pages={751-755},
}

@article{yanardag2015deep,
	title={Deep graph kernels},
	author={Yanardag, P. and Vishwanathan, S. V. N},
	journal={\textnormal{in} SIGKDD},
	pages={1365--1374},
	year={2015}
}

@inproceedings{shervashidze2009efficient,
	title={Efficient graphlet kernels for large graph comparison},
	author={Shervashidze, Nino and Vishwanathan, SVN and Petri, Tobias and Mehlhorn, Kurt and Borgwardt, Karsten},
	booktitle={Artificial intelligence and statistics},
	pages={488--495},
	year={2009},
	organization={PMLR}
}

@article{shervashidze2011weisfeiler,
	title={Weisfeiler-lehman graph kernels.},
	author={Shervashidze, Nino and Schweitzer, Pascal and Van Leeuwen, Erik Jan and Mehlhorn, Kurt and Borgwardt, Karsten M},
	journal={Journal of Machine Learning Research},
	volume={12},
	number={9},
	year={2011}
}

@inproceedings{zhang2018end,
	title={An End-to-End Deep Learning Architecture for Graph Classification},
	author={Zhang, Muhan and Cui, Zhicheng and Neumann, Marion and Chen, Yixin},
	booktitle={AAAI},
	pages={4438--4445},
	year={2018}
}

@article{kipf2016semi,
	title={Semi-supervised classification with graph convolutional networks},
	author={Kipf, Thomas N and Welling, Max},
	journal={arXiv preprint arXiv:1609.02907},
	year={2016}
}

@article{vaswani2017attention,
	title={Attention is all you need},
	author={Vaswani, Ashish and Shazeer, Noam and Parmar, Niki and Uszkoreit, Jakob and Jones, Llion and Gomez, Aidan N and Kaiser, {\L}ukasz and Polosukhin, Illia},
	journal={Advances in neural information processing systems},
	volume={30},
	year={2017}
}

@article{ying2021transformers,
	title={Do transformers really perform badly for graph representation?},
	author={Ying, Chengxuan and Cai, Tianle and Luo, Shengjie and Zheng, Shuxin and Ke, Guolin and He, Di and Shen, Yanming and Liu, Tie-Yan},
	journal={Advances in neural information processing systems},
	volume={34},
	pages={28877--28888},
	year={2021}
}

@article{wu2023sgformer,
	title={Sgformer: Simplifying and empowering transformers for large-graph representations},
	author={Wu, Qitian and Zhao, Wentao and Yang, Chenxiao and Zhang, Hengrui and Nie, Fan and Jiang, Haitian and Bian, Yatao and Yan, Junchi},
	journal={Advances in neural information processing systems},
	volume={36},
	pages={64753--64773},
	year={2023}
}

@article{li2012effective,
	title={Effective graph classification based on topological and label attributes},
	author={Li, Geng and Semerci, Murat and Yener, B{\"u}lent and Zaki, Mohammed J},
	journal={Statistical Analysis and Data Mining: The ASA Data Science Journal},
	volume={5},
	number={4},
	pages={265--283},
	year={2012},
	publisher={Wiley Online Library}
}

@article{debnath1991structure,
	title={Structure-activity relationship of mutagenic aromatic and heteroaromatic nitro compounds. correlation with molecular orbital energies and hydrophobicity},
	author={Debnath, Asim Kumar and Lopez de Compadre, Rosa L and Debnath, Gargi and Shusterman, Alan J and Hansch, Corwin},
	journal={Journal of medicinal chemistry},
	volume={34},
	number={2},
	pages={786--797},
	year={1991},
	publisher={ACS Publications}
}
\end{document}